%% file: main.tex
\documentclass[preprint]{article}

\usepackage{smile}
\usepackage[nonatbib]{arxiv}
\input{math_commands.tex}

\usepackage[utf8]{inputenc} 
\usepackage[T1]{fontenc}    
\usepackage{hyperref}       
\usepackage{url}            
\usepackage{booktabs}       
\usepackage{amsfonts}       
\usepackage{nicefrac}       
\usepackage{microtype}      
\usepackage{lipsum}
\usepackage{graphicx}
\usepackage{microtype}
\usepackage{subfigure}
\usepackage[shortlabels]{enumitem}
\usepackage{xcolor}
\usepackage{setspace}
\usepackage{float}
\usepackage{natbib}
\RequirePackage{natbib}

\setcitestyle{authoryear,round,citesep={;},aysep={,},yysep={;}}

\setcitestyle{square}
\setcitestyle{comma}
\setcitestyle{numbers}

\def\norm#1{\left\lVert#1\right\rVert}
\def\|#1\|{\norm{#1}}

\title{
Understanding Square Loss in Training Overparametrized Neural Network Classifiers
}

\author{Tianyang Hu \thanks{These authors contributed equally to this manuscript.} \\
  Huawei Noah’s Ark Lab\\
  \texttt{hutianyang1@huawei.com} \\
  \And
  Jun Wang $^*$ \\
  HKUST\\
  \texttt{jwangfx@connect.ust.hk} \\
  \AND
  Wenjia Wang $^*$\\
  HKUST \\
  \texttt{wenjiawang@ust.hk} \\
  \And
  Zhenguo Li \\
  Huawei Noah’s Ark Lab \\
  \texttt{li.zhenguo@huawei.com} \\
}

\begin{document}
\maketitle

\begin{abstract}
Deep learning has achieved many breakthroughs in modern classification tasks. 
Numerous architectures have been proposed for different data structures but when it comes to the loss function, the cross-entropy loss is the predominant choice. 
Recently, several alternative losses have seen revived interests for deep classifiers. In particular, empirical evidence seems to promote square loss but a theoretical justification is still lacking. 
In this work, we contribute to the theoretical understanding of square loss in classification by systematically investigating how it performs for overparametrized neural networks in the neural tangent kernel (NTK) regime.
Interesting properties regarding the generalization error, robustness, and calibration error are revealed. 
We consider two cases, according to whether classes are separable or not. 
In the general non-separable case, fast convergence rate is established for both misclassification rate and calibration error. 
When classes are separable, the misclassification rate improves to be exponentially fast. Further, the resulting margin is proven to be lower bounded away from zero, providing theoretical guarantees for robustness.  
We expect our findings to hold beyond the NTK regime and translate to practical settings. To this end, we conduct extensive empirical studies on practical neural networks, demonstrating the effectiveness of square loss in both synthetic low-dimensional data and real image data.  
Comparing to cross-entropy, square loss has comparable generalization error but noticeable advantages in robustness and model calibration. 
\end{abstract}

\section{introduction}
\label{intro}

The pursuit of better classifiers has fueled the progress of machine learning and deep learning research. The abundance of benchmark image datasets, e.g., MNIST, CIFAR, ImageNet, etc., provides test fields for all kinds of new classification models, especially those based on deep neural networks (DNN). 
With the introduction of CNN, ResNets, and transformers, DNN classifiers are constantly improving and catching up to the human-level performance. 
In contrast to the active innovations in model architecture, the training objective remains largely stagnant, with cross-entropy loss being the default choice.
Despite its popularity, cross-entropy has been shown to be problematic in some applications.
Among others, 
\citet{yu2020learning} argued that features learned from cross-entropy lack interpretability and proposed a new loss aiming for maximum coding rate reduction.
\citet{pang2019rethinking} linked the use of cross-entropy to adversarial vulnerability and proposed a new classification loss based on latent space matching. 
\citet{guo2017calibration} discovered that the confidence of most DNN classifiers trained with cross-entropy is not well-calibrated.

Recently, several alternative losses have seen revived interests for deep classifiers. In particular, many existing works have presented empirical evidence promoting the use of square loss over cross-entropy. 
\citet{hui2020evaluation} conducted large-scale experiments comparing the two and found that square loss tends to perform better in natural language processing related tasks while cross-entropy usually yields slightly better accuracy in image classification. 
Similar comparisons are also made in \citet{demirkaya2020exploring}.
\citet{kornblith2020demystifying} compared a variety of loss functions and output layer regularization strategies on the accuracy and out-of-distribution robustness, and found that square loss has greater class separation and better out-of-distribution robustness.

In comparison to the empirical investigation, theoretical understanding of square loss in training deep learning classifiers is still lacking. 
Through our lens, square loss has its uniqueness among classic classification losses, and we argue that it has great potentials for modern classification tasks. Below we list our motivations and reasons why.

\textbf{Explicit feature modeling}\quad
Deep learning's success can be largely attributed to its superior ability as feature extractors. For classification, the ideal features should be separated between classes and concentrated within classes. However, when optimizing cross-entropy loss, it's not obvious what the learned features should look like \citep{yu2020learning}. 
In the terminal stage of training, \cite{papyan2020prevalence} proved that when cross-entropy is sufficiently minimized, the penultimate layers features will collapse to the scaled simplex structure. Knowing this, would it be better to directly enforce the terminal solution by using square loss with the simplex coding \citep{mroueh2012multiclass}?
Unlike cross-entropy, square loss uses the label codings (one-hot, simplex etc.) as features, which can be modeled explicitly to control class separations.

\textbf{Model Calibration}\quad
An ideal classifier should not only give the correct class prediction, but also with the correct confidence.
Calibration error measures the closeness of the predicted confidence to the underlying conditional probability $\eta$. 
Using square loss in classification can be essentially viewed as regression where it treats discrete labels as continuous code vectors. It can be shown that the optimal classifier under square loss is $2\eta-1$, linear with the ground truth. This distinguishing property allows it to easily recover $\eta$. In comparison, the optimal classifiers under the hinge loss and cross-entropy are sign$(2\eta-1)$ and $\log(\frac{\eta}{1-\eta})$, respectively. Therefore, hinge loss doesn't provide reliable information on the prediction confidence, and cross-entropy can be problematic when $\eta$ is close to 0 or 1 \citep{zhang2004statistical}. Hence, in terms of model calibration, square loss is a natural choice. 

\textbf{Connections to popular approaches}\quad
Mixup \citep{zhang2017mixup} is a popular data augmentation technique where augmented data are constructed via convex combinations of inputs and their labels. Like in square loss, mixup treats labels as continuous and is shown to improve the generalization of DNN classifiers. 
In knowledge distillation \citep{hinton2015distilling}, where a student classifier is trying to learn from a trained teacher, \citet{menon2021statistical} proved that the ``optimal" teacher with the ground truth conditional probabilities provides the lowest variance in student learning. Since classifiers trained using square loss is a natural consistent estimator of $\eta$, one can argue that it is a better teacher. 
In supervised contrastive learning \citep{khosla2020supervised}, the optimal features are the same as those from square loss with simplex label coding \citep{graf2021dissecting} (details in Section \ref{sec:encoding}). %

Despite its lack of popularity in practice, square loss has many advantages that can be easily overlooked. 
In this work, we systematically investigate from a statistical estimation perspective, the properties of deep learning classifiers trained using square loss. 
The neural networks in our analysis are required to be sufficiently overparametrized in the neural tangent kernel (NTK) regime. 
Even though this restricts the implication of our results, it is a necessary first step towards a deeper understanding.
In summary, our main contributions are: 
\begin{itemize}
    \item
    \underline{Generalization error bound}: We consider two cases, according to whether classes are separable or not. In the general non-separable case, we adopt the classical binary classification setting with smooth conditional probability. Fast rate of convergence is established for overparametrized neural network classifiers with Tsybakov's noise condition. If two classes are separable with positive margins, we show that overparametrized neural network classifiers can provably reach zero misclassification error with probability \textit{exponentially} tending to one. To the best of our knowledge, this is the \textit{first} such result for separable but not linear separable classes. Furthermore, we bridge these two cases and offer a \textit{unified} view by considering auxiliary random noise injection. 
    
    \item
     \underline{Robustness (margin property)}: When two classes are separable, the decision boundary is not unique and large-margin classifiers are preferred. 
     In the separable case, we further show that the decision boundary of overparametrized neural network classifiers trained by square loss cannot be too close to the data support and the resulting margin is lower bounded away from zero, providing theoretical guarantees for robustness. 
    \item
    \underline{Calibration error}: 
    We show that classifiers trained using square loss are inherently well-calibrated, i.e., the trained classifier provides consistent estimation of the ground-truth conditional probability in $L_\infty$ norm.  Such property doesn't hold for cross-entropy. 
  
    \item
    \underline{Empirical evaluation}: We corroborate our theoretical findings with empirical experiments in both synthetic low-dimensional data and real image data. Comparing to cross-entropy, square loss has comparable generalization error but noticeable advantages in robustness and model calibration.
    
\end{itemize} 

This work contributes to the theoretical understanding of deep classifiers, from a nonparametric estimation point of view, which has been a classic topic in statistics literature. 
Among others, \citet{mammen1999smooth} established the optimal convergence rate for 0-1 loss excess risk when the decision boundary is smooth.
\citet{zhang2004statistical,bartlett2006convexity} extended the analysis to various surrogate losses. 
\citet{audibert2007fast, kohler2007rate} studied the convergence rates for plug-in classifiers from local averaging estimators. 
\citet{steinwart2007fast} investigated the convergence rate for support vector machine using Gaussian kernels. 
We build on and extend classic results to neural networks in the NTK regime. 
There are existing nonparametric results on deep classifiers, e.g., \citet{kim2018fast} derived fast convergence rates of DNN classifiers that minimize the empirical hinge loss, 
\cite{kohler2020statistical} considered a family of CNN-inspired classifiers and developed similar results for the empirical cross-entropy minimizer, etc. 
Unlike aforementioned works that only concern the existence of a good classifier (with theoretical worst-case guarantee), in ignorance of the tremendous difficulty of neural network optimization, 
our results further incorporate the training algorithm and apply to trained classifiers, which relates better to practice. To the best of the authors' knowledge, 
similar attainable fast rates (faster than $n^{-\frac{1}{2}}$) have never been established for neural network classifiers.

We require the neural network to be overparametrized, which has been extensively studied recently, under the umbrella term NTK. Most such results are in the regression setting with a handful of exceptions. 
\citet{ji2019polylogarithmic} showed that only polylogarithmic width is sufficient for gradient descent to overfit the training data using logistic loss.  
\citet{hu2020simple} proved generalization error bound for regularized NTK in classification.  
\citet{cao2019generalization, cao2020generalization}
provided optimization and generalization guarantees for overparametrized network trained with cross-entropy. 
In comparison, our results are sharper in the sense that we take the ground truth data assumptions into consideration. This allows a faster convergence rate, especially when the classes are separable, where the exponential convergence rate is attainable.
The NTK framework greatly reduces the technical difficulty for our theoretical analysis. However, our results are mainly due to properties of the square loss itself and we expect them to hold for a wide range of classifiers. 

There are other works investigating the use of square loss for training (deep) classifiers.
\citet{han2021neural} uncovered that the ``neural collapse" phenomenon also occurs under square loss where the last-layer
features eventually collapse to their simplex-style class-means. 
\citet{muthukumar2020classification} compared classification and regression tasks in the overparameterized linear model with Gaussian features, illustrating different roles and properties of loss functions used at the training and testing phases. 
\citet{poggio2019generalization} made interesting observations on effects of popular regularization techniques such as batch normalization and weight decay on the gradient flow dynamics under square loss. 
These findings support our theoretical results' implication, which further strengthens our beliefs that the essence comes from the square loss and our analysis can go beyond NTK regime.

The rest of this paper is arranged as follows. Section \ref{pre} presents some preliminaries. Main theoretical results are in Section \ref{sec:theory}. The simplex label coding is discussed in Section \ref{sec:encoding} followed by numerical studies in Section \ref{simulation} and conclusions in Section \ref{sec:conclude}. Technical proofs and details of the numerical studies can be found in the Appendix.

\section{Preliminaries}\label{pre}

\textbf{Notation}\quad
For a function $f:\Omega\to\mathbb{R}$, let
$\|f\|_\infty=\sup_{\bx\in\Omega}|f(\bx)|$ and $\|f\|_p=(\int_{\Omega} |f(\bx)|^p d\bx)^{1/p}$. 
For a vector $\bx$, $\|\bx\|_p$ denotes its $p$-norm, for $1\leq p \leq \infty$.
$L_p$ and $l_p$ are used to distinguish function norms and vector norms. For two positive sequences $\{a_n\}_{n\in \mathbb{N}}$ and $\{b_n\}_{n\in \mathbb{N}}$, we write $a_n\lesssim b_n$ if there exists a constant $C>0$ such that $a_n\leq C b_n$ for all sufficiently large $n$. We write $a_n\asymp b_n$ if $a_n\lesssim b_n$ and $b_n\lesssim a_n$. Let $[N]=\{1,\dots, N\}$ for $N\in\mathbb{N}$, $\mathbb{I}$ be the indicator function, and $\bI_d$ be the $d\times d$ identity matrix. $N(\mu, \mathbf{\Sigma})$ represents Gaussian distribution with mean $\mu$ and covariance $\mathbf{\Sigma}$.

\textbf{Classification problem settings}\quad Let $P$ be an underlying probability measure on $\Omega \times \bY$, where $\Omega \subset \mathbb{R}^{d}$ is compact and ${\bY} = \{1,-1\}$. Let $(X,Y)$ be a random variable with respect to $P$. Suppose we have observations $\{(\bx_i,y_i)\}_{i=1}^{n} \subset (\Omega \times Y)^{n}$ i.i.d. sampled according to $P$. The classification task is to predict the unobserved label $y$ given a new input $\bx \in \Omega$. Let $\eta$ defined on $\Omega$ denote the conditional probability, i.e., $\eta(\bx)=\mathbb{P}(y=1|\bx)$. Let $P_{X}$ be the marginal distribution of $P$ on $X$.
The key quantity of interest is the misclassification error, i.e., 0-1 loss. In the population level, the 0-1 loss can be written as  
\begin{align}\label{bayesrisk}
    L(f) = \mathbb{E}_{(X,Y)\sim P}\mathbb{I}\{{\rm sign}(f(X))\neq Y\} = & \mathbb{E}_{X\sim P_X}[(1-\eta(X))\mathbb{I}\{f(X)\geq 0\} + \eta(X)\mathbb{I}\{f(X)< 0\}],
\end{align}
where the expectation is taken with respect to the probability measure $P$. Clearly, an optimal classifier with the minimal 0-1 loss is $2\eta-1$. 

According to whether labels are deterministic, there are two scenarios of interest. 
If $\eta$ only takes values from $\{0,1\}$, i.e., labels are deterministic, we call this case the \textit{separable case}\footnote{In the separable case we consider, the classes are not limited to linearly separable but can be arbitrarily complicated.}. Let $\Omega_1=\{\bx|\eta(\bx)=1\}$, $\Omega_2=\{\bx|\eta(\bx)=0\}$ and $\Omega=\Omega_1\cup \Omega_2$. 
If the probability measure of $\{\bx|\eta(\bx)\in(0,1)\}$ is non-zero, i.e., the labels contain randomness, we call this case the \textit{non-separable case}.
In the separable case, we further assume that there exists a positive margin, i.e., dist$(\Omega_1,\Omega_2)\geq 2\gamma >0$, where $\gamma$ is a constant, and dist$(\Omega_1,\Omega_2)=\inf_{\bx\in \Omega_1,\bx'\in \Omega_2}\|\bx-\bx'\|_2$.
In the non-separable case, to quantify the difficulty of classification, we adopt the well-established Tsybakov's noise condition \citep{audibert2007fast}, which measures how large the ``difficult region'' is where $\eta(\bx)\approx 1/2$.

\begin{definition}[Tsybakov's noise condition]
\label{def:tsybakov}
Let $\kappa\in [0,\infty]$. We say $P$ has Tsybakov noise exponent $\kappa$ if there exists a constant $C,T > 0$ such that for all 
 $0<t<T$,
$
    P_{X}(|2\eta(X) - 1| < t) \le C \cdot t^{\kappa}.
$
\end{definition}
A large value of $\kappa$ implies the difficult region to be small. It is expected that a larger $\kappa$ leads to a faster convergence rate of a neural network classifier. This intuition is verified for the overparametrized neural network classifier trained by square loss and $\ell_2$ regularization. See Section \ref{sec:theory} for more details.

\textbf{Neural network setup}\quad We mainly focus on the one-hidden-layer ReLU neural network family $\mathcal{F}$ with $m$ nodes in the hidden layer, denoted by \[f_{\bW,\ba}(\bx) = \frac{1}{\sqrt{m}}\sum_{r=1}^{m} a_r \sigma(\bW_r^\top \bx),\]
where $\bx\in \Omega$, $\bW=(\bW_1,\cdots,\bW_m)\in \mathbb{R}^{d\times m}$ is the weight matrix in the hidden layer, $\ba=(a_1,\cdots,a_m)^\top\in \mathbb{R}^m$ is the weight vector in the output layer, $\sigma(z) = \max\{0,z\}$ is the rectified linear unit (ReLU). The initial values of the weights are independently generated from 
\begin{align*}
    \bW_r(0) ~\sim~ N(\mathbf{0},\xi^2\bI_m),~~a_r~\sim~ \mathrm{unif}\{-1, 1\},~~\forall r\in [m].
\end{align*}
Based on the observations $\{(\bx_i,y_i)\}_{i=1}^{n}$, the goal of training a neural network is to find a solution to
\begin{align}\label{eq:optgoal}
    \min_{\bW} \sum_{i=1}^n l(f_{\bW,\ba}(\bx_i),y_i) + \mu\mathcal{R}(\bW,\ba),
\end{align}
where $l$ is the loss function, $\mathcal{R}$ is the regularization, and $\mu\geq 0$ is the regularization parameter. Note in \Eqref{eq:optgoal} that we only consider training the weights $\bW$. This is because $a\cdot\sigma(z) = \mbox{sign}(a)\cdot\sigma(|a|z)$, which allows us to reparametrize the network to have all $a_i$'s to be either $1$ or $-1$. In this work, we consider square loss associated with $\ell_2$ regularization, i.e., $l(f_{\bW,\ba}(\bx_i),y_i) = (f_{\bW,\ba}(\bx_i)-y_i)^2$ and $\mathcal{R}(\bW,\ba)=\|\bW\|_2^2$.

A popular way to train the neural network is via gradient based methods. It has been shown that the training process of DNNs can be characterized by the neural tangent kernel (NTK) \citep{jacot2018neural}. As is usually assumed in the NTK literature \citep{arora2019fine, hu2020simple, bietti2019inductive,hu2021regularization}, we consider data on the unit sphere $\mathbb{S}^{d-1}$, i.e., $\|\bx_i\|_2=1,\forall i\in [n]$, and the neural network is highly overparametrized ($m\gg n$) and trained by gradient descent (GD). For details about NTK and GD in one-hidden-layer ReLU neural networks, we refer to Appendix \ref{app:GDNTK}.  In the rest of this work, we use $f_{\bW(k),\ba}$ to denote the GD-trained neural network classifier under square loss associated with $\ell_2$ regularization, where $k$ is the iteration number satisfying Assumption \ref{as1} and $\bW(k)$ is the weight matrix after $k$-th iteration.

\section{Theoretical Results}\label{sec:theory}

In this section, we present our main theoretical results. Throughout the analysis, we assume that the overparametrized neural network $f_{\bW,\ba}$ and the training process via GD satisfy Assumption \ref{as1} (see Appendix \ref{app:assumptions}), which essentially requires the neural network to be sufficiently overparametrized (with a finite width), and imposes conditions on the learning rate and iteration number. Our theoretical results consist of three parts: generalization error, robustness, and calibration error.

\subsection{Generalization error bound}\label{subsec:acc}
In classification, the generalization error is typically referred to as the misclassification error, which can be quantified by $L(f)$ defined in \Eqref{bayesrisk}. In the non-separable case, the excess risk, defined by $L(f)-L^*$, is used to evaluate the quality of a classifier $f$, where $L^* = L(2\eta-1)$, which minimizes the 0-1 loss.
The following theorem states that the overparametrized neural network with GD and $\ell_2$ regularization can achieve a small excess risk in the non-separable case.

\begin{theorem}[Excess risk in the non-separable case]\label{thm:nonsepthm1}
Suppose Assumptions \ref{as1}, \ref{as2}, and \ref{as6} hold. Assume the conditional probability $\eta(\bx)$ satisfies Tsybakov's noise condition with component $\kappa$. Let $\mu \asymp n^{\frac{d-1}{2d-1}}$. Then 
\begin{equation}\label{eq:thmnons1x}
    L(f_{\bW(k),\ba}) = L^{*} +  O_{\mathbb{P}}(n^{-\frac{d(\kappa+1)}{(2d-1)(\kappa + 2)}}).
\end{equation}
\end{theorem}
From Theorem \ref{thm:nonsepthm1}, we can see that as $\kappa$ becomes larger, the convergence rate becomes faster, which is intuitively true. Generalization error bounds in this setting is scarce. To the best of the authors' knowledge,  \citet{hu2020simple} is the closest work (the labels are randomly flipped), where the bound is in the order of $O_\PP(1/\sqrt{n})$. Our bound is faster, especially with larger $\kappa$. It is known that the optimal convergence rate under Assumptions \ref{as2} and \ref{as6} is $O_{\mathbb{P}}(n^{-\frac{d(\kappa+1)}{d\kappa+4d-2}})$ \citep{audibert2007fast}. The differences between \Eqref{eq:thmnons1x} and the optimal convergence rate is that there is an extra $(d-1)\kappa$ in the denominator of the convergence rate in \Eqref{eq:thmnons1x} (since $n^{-\frac{d(\kappa+1)}{(2d-1)(\kappa + 2)}}=n^{-\frac{d(\kappa+1)}{(d-1)\kappa + d\kappa+4d-2}}$). If the conditional probability $\eta$ has a bounded Lipschitz constant, then \citet{kohler2007rate} showed  that the convergence rate based on the plug-in kernel estimate is $O_{\mathbb{P}}(n^{-\frac{\kappa+1}{\kappa+3+d}})$, which is slower than the rate in \Eqref{eq:thmnons1x} if $d$ is large.

Now we turn to the separable case. Since $\eta$ only takes value from $\{0,1\}$ in the separable case, $\eta$ is bounded away from 1/2. Therefore, one can trivially take $\kappa\rightarrow \infty$ in \Eqref{eq:thmnons1x} and obtain the convergence rate $O_{\mathbb{P}}(n^{-d/(2d-1)})$. However, this rate can be significantly improved in the separable case, as stated in the following theorem.

\begin{theorem}[Generalization error in the separable case]\label{thm:sep}
Suppose Assumptions \ref{as1}, \ref{as4}, and \ref{as5} hold. Let $\mu =o(1)$. There exist positive constants $C_1,C_2$ such that the misclassification rate is 0\% with probability at least $1-\delta - C_1\exp(-C_2n)$, and $\delta$ can be arbitrarily small\footnote{The term $\delta$ only depends on the width of the neural network. A smaller $\delta$ requires a wider neural network. If $\delta=0$, then the number of nodes in the hidden layer is infinity.} by enlarging the neural network's width.
\end{theorem}

Note that in Theorem \ref{thm:sep}, the regularization parameter can take any rate that converges to zero. In particular, $\mu$ can be zero, and the corresponding classifier overfits the training data. Theorem \ref{thm:sep} states that the convergence rate in the separable case is exponential, if a sufficiently wide neural network is applied. This is because the observed labels are not corrupted by noise, i.e., $\PP(y=1|\bx)$ is either one or zero. Therefore, it is easier to classify separable data, which is intuitively true.

\subsection{Robustness and calibration error}\label{subsec:robust}
If two classes are separable with positive margin, the decision boundary is not unique. Practitioners often prefer the decision boundary with large margins, which are robust against possible perturbation on input points \citep{elsayed2018large, ding2018mma}. The following theorem states that the square loss trained margin can be lower bounded by a positive constant. 
Recall that in the separable case, $\Omega=\Omega_1\cup \Omega_2$, where $\Omega_1=\{\bx|\eta(\bx)=1\}$ and $\Omega_2=\{\bx|\eta(\bx)=0\}$. 
\begin{theorem}[Robustness in the separable case]\label{prop:sep}
Suppose the assumptions of Theorem \ref{thm:sep} are satisfied. Let $\mu =o(1)$. Then there exist positive constants $C,C_1,C_2$ such that for all $n$,
\begin{align*}
    \min_{\vx\in \mathcal{D}_T, \vx'\in \Omega_1\cup \Omega_2}\|\vx-\vx'\|_2 \geq C,
\end{align*}
and the misclassification rate is 0\% with probability at least $1-\delta-C_1\exp(-C_2n)$, where $\mathcal{D}_T$ is the decision boundary, and $\delta$ is as in Theorem \ref{thm:sep}.

\end{theorem}
\begin{remark}
Note that $\|\vx-\vx'\|_\infty\geq \sqrt{d}\|\vx-\vx'\|_2$, thus Theorem \ref{prop:sep} also indicates $l_\infty$ robustness.
\end{remark}

In the non-separable case, $\eta(\bx)$ varies within (0,1) and practitioners may not only want a classifier with a small excess risk, but also want to recover the underlying conditional probability $\eta$. Therefore, square loss is naturally preferred since it treats the classification problem as a regression problem. The following theorem states that, one can recover the conditional probability $\eta$ by using an overparametrized neural network with $\ell_2$ regularization and GD training.

\begin{theorem}[Calibration error]\label{thm:nonsepcali}
Suppose the conditions in Theorem \ref{thm:nonsepthm1}, Assumption \ref{as4} and \ref{as6} are fulfilled. 
Let $\mu \asymp n^{\frac{d-1}{2d-1}}$. Then
\begin{align}
    \|(f_{\bW(k),\ba} + 1)/2-\eta\|_{L_{\infty}} = O_{\mathbb{P}}(n^{-\frac{1}{4d-2}}).
\end{align}
\end{theorem}
Theorem \ref{thm:nonsepcali} states that the underlying conditional probability in the non-separable case can be recovered by $(f_{\bW(k),\ba}+1)/2$. The form $(f_{\bW(k),\ba}+1)/2$ is to account for the $\{-1,1\}$ label coding. Under \{0,1\} coding, the estimator would be $f_{\bW(k),\ba}$ itself. 
The $L_\infty$ consistency doesn't hold for cross-entropy trained neural networks, due to the form of the optimal solution $\log(\frac{\eta}{1-\eta})$. With limited capacity, the network's confidence prediction is bounded away from 0 and 1 \citep{zhang2004statistical}. 
In practice, we want to control the complexity of the neural network thus it is usually the case that $\|f_{\bW(k),\ba}\|_\infty < C$ for some constant $C$.
Hence, it cannot accurately estimate $\eta(\bx)$ when $\eta(\bx)>\frac{e^C}{1+e^C}$ or $\eta(\bx) < \frac{1}{1+e^C}$, which makes the calibration error under the cross-entropy loss always bounded away from zero. However, square loss does not have such a problem.

Notice that the calibration error bound in Theorem \ref{thm:nonsepcali} does not depend on the Tsybakov's noise condition, and is slower than the excess risk. This is because, a small calibration error is much stronger than a small excess risk, since the former requires the conditional probability estimation to be \textit{uniformly} accurate, not just matching the sign of $\eta(\bx)-1/2$. To be more specific, a good estimated $\hat{\eta}$ can always result in a low risk plug-in classifier $\hat{f}(\bx)=2\hat{\eta}(\bx)-1$, but not vice versa.

\begin{remark}[Technical challenge]
Despite the similar forms of regression and classification using square loss, most of the regression analysis techniques cannot be directly applied to the classification problem, even if the supports of two classes are non-separable. Moreover, it is clear that classification problems in the separable case are completely different with regression problems. 
\end{remark}

\begin{remark}[Extension on NTK]
Although our analysis only concerns overparametrized one-hidden-layer ReLU neural networks, it can potentially apply to other types of neural networks in the NTK regime. Recently, it has been shown that overparametrized multi-layer networks correspond to the Laplace kernel \citep{geifman2020similarity,chen2020deep}. 
As long as the trained neural networks can approximate the classifier induced by the NTK, our results can be naturally extended.
\end{remark}

\subsection{Transition from separable to non-separable}
The general non-separable case and the special separable case can be connected via Gaussian noise injection. 
In practice, data augmentation is an effective way to improve robustness and the simplest way is Gaussian noise injection \citep{he2019parametric}. In this section, we only consider it as an auxiliary tool for theoretical analysis purpose and not for actual robust training.
Injecting Gaussian noise amounts to convoluting a Gaussian distribution $N(0,\upsilon^2 \bI_d)$ to the marginal distribution $P_X$, which enlarges both $\Omega_1$ and $\Omega_2$ to $\mathbb{R}^d$ and a unique decision boundary $\mathcal{D}_\upsilon$ can be induced. 
Correspondingly, the ``noisy" conditional probability, denoted as $\tilde{\eta}_\upsilon$, is also smoothed to be continuous on $\mathbb{R}^d$. As $\upsilon\to 0$, $\|\tilde{\eta}_\upsilon-\eta\|_\infty\to 0$ on $\Omega_1$ and $\Omega_2$ and the limiting $\tilde{\eta}_0$ is a piecewise constant function with discontinuity at the induced decision boundary. 

\begin{lemma}[Tsybakov's noise condition under Gaussian noises]
\label{lemma:noisy_tsybakov}
Let the margin be $ 2\gamma>0$, the noise be $N(0, \upsilon^2 \bI_d)$. Then there exist some constants $T, C>0$ such that for any $0<t<T$,
 \begin{equation*}
     P_{X}(|2\tilde{\eta}_\upsilon(X) - 1| < t) \le \frac{C\upsilon^2}{\gamma}\exp\left({-\frac{\gamma^2}{2\upsilon^2}}\right) \cdot t.
 \end{equation*}
\end{lemma}

\begin{theorem}[Exponential convergence rate]
\label{thm:noise}
Suppose the classes are separable with margin $2\gamma>0$. No matter how complicated $\Omega_1\cup \Omega_2$ are, the excess risk of the over parameterized neural network classifier satisfying Assumptions \ref{as1} and \ref{as6} has the rate $O_\PP(e^{-n\gamma /7})$.
\end{theorem}
The proof of Theorem \ref{thm:noise} involves taking the auxiliary noise to zero, e.g., $v=v_n\asymp 1/\sqrt{n}$. The exponential convergence rate is a direct outcome of Lemma \ref{lemma:noisy_tsybakov} and Theorem \ref{thm:nonsepthm1}. Note that our exponential convergence rate is much faster than existing ones under the similar separable setting \citep{ji2019polylogarithmic, cao2019generalization, cao2020generalization}, which are all polynomial with $n$, e.g., $O_\PP(1/\sqrt{n})$. 

\begin{remark}
Theorems \ref{thm:nonsepcali} and \ref{thm:noise} share the same gist that the over parameterized neural network classifiers can have exponential convergence rate when data are separable with positive margin, while the result of Theorem \ref{thm:noise} is weaker than that of Theorem \ref{thm:nonsepcali}, but with milder conditions. Nevertheless, Theorem \ref{thm:noise} bridges the non-separable case and separable case.
\end{remark}

\section{Multiclass Classification}
\label{sec:encoding}
In binary classification, the labels are usually encoded as $-1$ and $1$. 
When there are $K>2$ classes, the default label coding is one-hot.
However, it is empirically observed that this vanilla square loss struggles when the number of classes are large, for which scaling tricks have been proposed \citep{hui2020evaluation, demirkaya2020exploring}. 
Another popular coding scheme is the simplex coding \citep{mroueh2012multiclass}, which takes maximally separated $K$ points on the sphere as label features. When $K=2$, this reduces to the typical $-1,1$ coding.
Many advantages of the simplex coding have been discussed, including its relationship with cross-entropy loss and supervised contrastive learning \citep{papyan2020prevalence, han2021neural, graf2021dissecting, fang2021exploring}.

In this work, we adopt the simplex coding. More discussion and empirical comparison about the coding choices can be found in Appendix \ref{app:realdata}. Given the label coding, one can easily generalize the theoretical development in Section \ref{sec:theory} by employing the following objective function
\begin{align*}
    \min_{\bW} \sum_{j=1}^K\sum_{i=1}^n (f_{j,\bW,\ba}(\bx_i)-y_{i,j})^2 + \mu\|\bW\|_2^2,
\end{align*}
where $f_{\bW,\ba}:\Omega \mapsto \mathbb{R}^K$, and $\by_i=(y_{i,1},...,y_{i,K})^\top$ is the label of $i$-th observation.

The following proposition states a relationship between the simplex coding scheme and the conditional probability.

\begin{proposition}[Conditional probability]
\label{prop:labeltr}
Let $f^*:\Omega \to \RR^K$ minimize the mean square error
$
    \mathbb{E}_X(f^*(X)-\bv_y)^2,
$
where $\bv_y$ is the simplex coding vector of label $y$. Then we have
\begin{align}
\label{eqn:calibration}
    \eta_k(\bx) := \mathbb{P}\left(y=k|\bx\right) = \left((K-1)f^*(\bx)^\top\bv_k+1\right)/K.
\end{align}
\end{proposition}
Unlike the softmax function when using cross entropy, the estimated conditional probability using square loss is not guaranteed to be within 0 and 1. This will cause issues for adversarial attacks, which will be discussed in detail in Appendix \ref{app:realdata}.

\section{NUMERICAL STUDIES}
\label{simulation}
Although our theoretical results are for overparametrized neural network in the NTK regime, we expect our conclusions to generalize to practical network architectures. The focus of this section is not on improving the state-of-the-art performance for deep classifiers, but to illustrate the difference between cross-entropy and square loss.
We provide experiment results on both synthetic and real data, to support our theoretical findings and illustrate the practical benefits of square loss in training overparametrized DNN classifiers. Compared with cross-entropy, the square loss has comparable generalization performance, but with stronger robustness and smaller calibration error. 

\subsection{Synthetic Data}  
We consider the square loss based and cross-entropy based overparametrized neural networks (ONN) with $\ell_2$ regularization, denoted as SL-ONN + $\ell_2$ and CE-ONN + $\ell_2$, respectively. The chosen ONNs are two-hidden-layer ReLU neural networks with 500 neurons for each layer, and the parameter $\mu$ is selected via a validation set. More implementation details are in Appendix \ref{app:synthetic}.
 
\begin{figure}[h]
    \centering
    \includegraphics[width=1\textwidth]{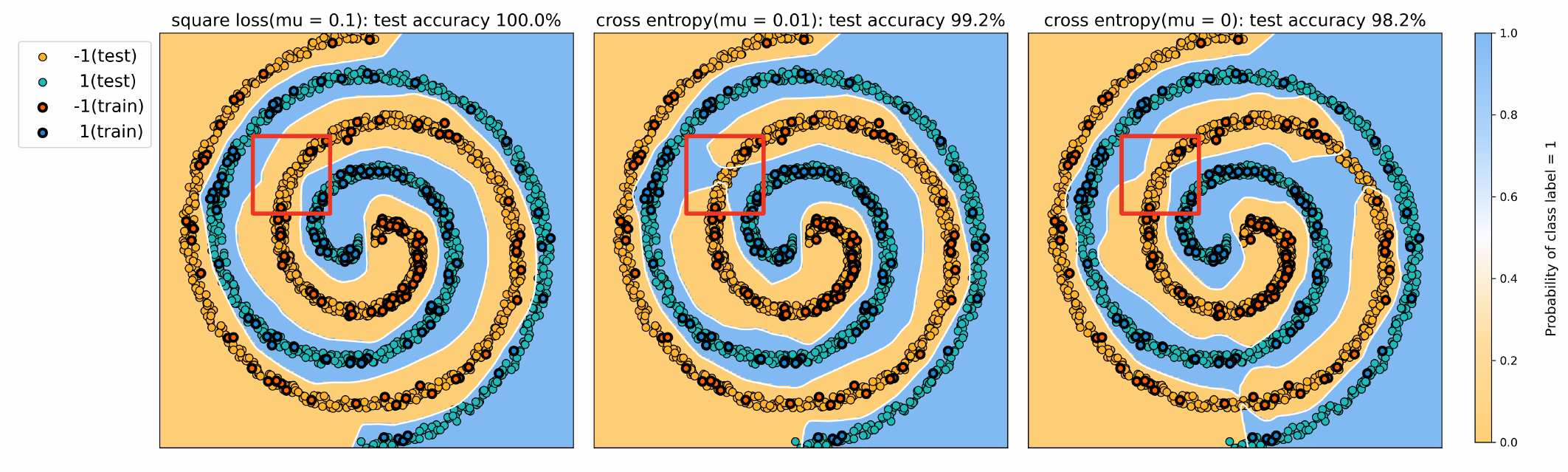}
    \vspace{-8mm}
    \caption{Test misclassification rates and decision boundaries predicted by: SL-ONN + $\ell_2$ (Left);  CE-ONN + $\ell_2$ (Center);   CE-ONN (Right) for the separable case. }
    \label{fig:sep}
    \vspace{-4mm}
\end{figure}

\textbf{Separable case}\quad We consider two separated classes with spiral curve like supports. 
We also present the performance of the cross-entropy based ONN without $\ell_2$ regularization (CE-ONN). 
Figure \ref{fig:sep} shows one instance of the test misclassification rate and decision boundaries attained by SL-ONN + $\ell_2$ (Left), CE-ONN + $\ell_2$ (Center), and CE-ONN (Right). 
From this example and other examples in Appendix \ref{app:synthetic}, it can be seen that SL-ONN + $\ell_2$ has a smaller test misclassification rate and a much smoother decision boundary. In particular, in the red region, where the training data are sparse, SL-ONN + $\ell_2$ fits the correct data distribution best. 

\textbf{Non-separable case}\quad We consider the conditional probability $\eta(\boldsymbol{x}) = \sin(\sqrt{2}\pi\vert|\boldsymbol{x}\vert|_{2}),\bx\in [-1,1]^2$, and the calibration performance of SL-ONN + $\ell_2$ and CE-ONN + $\ell_2$, where the classifiers are denoted by $\hat{f}_{l2}$ and $\hat{f}_{ce}$, respectively. 
The results are presented in Figure \ref{fig:non_sep} in the Appendix. The error bar plot of the test calibration error shows that $\hat{f}_{l2}$ has the smaller mean and standard deviation than $\hat{f}_{ce}$. It suggests that square loss generally outperforms cross entropy in calibration. The histogram and kernel density estimation of the test calibration errors for one case 
show that the pointwise calibration errors on the test points of $\hat{f}_{l2}$ are more concentrated around zero than those of $\hat{f}_{ce}$. Moreover, despite a comparable misclassification rate with $\hat{f}_{ce}$, $\hat{f}_{l2}$ has a smaller calibration error. Figure \ref{fig:non_sep} demonstrates that SL-ONN + $\ell_2$ recovers $\eta$ much better than CE-ONN + $\ell_2$.

\subsection{Real Data}
To make a fair comparison, we adopt popular architectures, ResNet \citep{he2016deep} and Wide ResNet \citep{zagoruyko2016wide} and evaluate them on the CIFAR image classification datasets, with only the training loss function changed, from cross-entropy (CE) to square loss with simplex coding (SL). 
Further, we don't employ any large scale hyper-parameter tuning and all the parameters are kept as default except for the learning rate  (lr) and batch size (bs), where we are choosing from the better of (lr=0.01, bs=32) and (lr=0.1, bs=128). 
Each experiment setting is replicated 5 times and we report the average performance followed by its standard deviation in the parenthesis. (lr=0.01, bs=32) works better for the most cases except for square loss trained WRN-16-10 on CIFAR-100. 
More experiment details and additional results can be found in Appendix \ref{app:realdata}.

\textbf{Generalization}\quad
In both CIFAR-10 and CIFAR-100, the performance of cross-entropy and square loss with simplex coding are quite comparable, as observed in \cite{hui2020evaluation}. 
Cross-entropy tends to perform slightly better for ResNet, especially on CIFAR-100 with an advantage of less than 1\%. 
There is a more significant gap with Wide ResNet where square loss outperforms cross-entropy by more than 1\% on both CIFAR-10 and CIFAR-100. The details can be found in Table \ref{table1}. 
\input{tables/table1}

\textbf{Adversarial robustness}\quad
Naturally trained deep classifiers are found to be adversarially vulnerable and adversarial attacks provide a powerful tool to evaluate classification robustness. For our experiment, we consider the black-box Gaussian noise attack, the classic white-box PGD attack \citep{madry2017towards} and the state-of-the-art AutoAttack \citep{croce2020reliable}, with attack strength level 2/255, 4/255, 8/255 in $l_\infty$ norm. AutoAttack contains both white-box and black-box attacks and offers a more comprehensive evaluation of adversarial robustness. 
The Gaussian noises results are presented in Table \ref{table:gaussian} in the Appendix. At different noise levels, square loss consistently outperforms cross-entropy, especially for WRN-16-10, with around 2-4\% accuracy improvement. More details can be found in Appendix \ref{app:realdata}.
The PGD and AutoAttack results are reported in Table \ref{table1}. Even though classifiers trained with square loss is far away from adversarially robust, it consistently gives significantly higher adversarial accuracy. 
The same margin can be carried over to standard adversarial training as well. Table \ref{table_adv} lists results from standard PGD adversarial training with CE and SL. By substituting cross-entropy loss to square loss, the robust accuracy increased around 3\% while maintaining higher clean accuracy. 

One thing to notice is that when constructing white-box attacks, square loss will not work well since it doesn't directly reflect the classification accuracy. More specifically, for a correctly classified image $(\bx,y)$, maximizing the square loss may result in linear scaling of the classifier $f(\bx)$, which doesn't change the predicted class (see Appendix \ref{app:realdata} for more discussion). 
To this end, we consider a special attack for classifiers trained by square loss by maximizing the cosine similarity between $f(\bx)$ and $\bv_y$. We call this angle attack and also utilize it for the PGD adversarial training paired with square loss in Table \ref{table_adv}.
In our experiments, this special attack rarely outperforms the standard PGD with cross-entropy and the reported PGD accuracy are from the latter settings. 
This property of square loss may be an advantage in defending adversarial attacks. 

\begin{figure}[ht]
    \centering
    \includegraphics[width=1\textwidth]{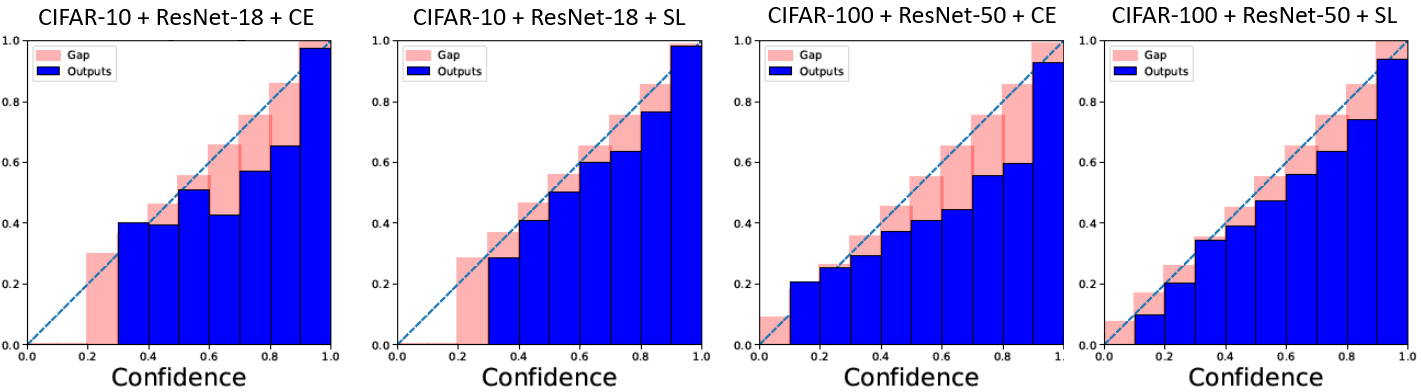}
    \caption{Reliability diagrams of ResNet-18 on CIFAR-10 and ResNet-50 on CIFAR-100. Square loss trained models behave more well-calibrated while cross-entropy trained ones tend to be visibly more over-confident.}
    \label{fig:ece}
\end{figure}

\vspace{-2mm}
\input{tables/table_add1}
\textbf{Model calibration}\quad
The predicted class probabilities for square loss  can be obtained from \Eqref{eqn:calibration}. 
Expected calibration error (ECE) measures the absolute difference between predicted confidence and  actual accuracy. Deep classifiers are usually found to be over-confident \citep{vaicenavicius2019evaluating}. 
Using ResNet as an example, we report the typical reliability diagram in Figure \ref{fig:ece}.
On CIFAR-10 with ResNet-18,
the average ECE for cross-entropy is 0.028 (0.002) while that for square loss is 0.0097 (0.001).
On CIFAR-100 with ResNet-50,
the average ECE for cross-entropy is 0.094 (0.005) while that for square loss is 0.068 (0.005).
Square loss results are much more calibrated with significantly smaller ECE. 

\section{Conclusions}
\label{sec:conclude}
Classification problems are ubiquitous in deep learning. As a fundamental problem, any progress in classification can potentially benefit numerous relevant tasks. 
Despite its lack of popularity in practice, square loss has many advantages that can be easily overlooked. 
Through both theoretical analysis and empirical studies, we identify several ideal properties of using square loss in training neural network classifiers, including provable fast convergence rates, strong robustness, and small calibration error. 
We encourage readers to try square loss in your own application scenarios.

\bibliography{references}

\appendix
\newpage
\setcounter{equation}{0}
\setcounter{page}{1}
\renewcommand{\thetable}{\Alph{section}.\arabic{table}}
\renewcommand{\thefigure}{\Alph{section}.\arabic{figure}}

\section{Gradient Descent and Neural Tangent Kernel}\label{app:GDNTK}

\paragraph{Gradient Descent}
Since we consider the square loss and $\ell_2$ regularization, the optimization problem in \Eqref{eq:optgoal} becomes
\begin{align}\label{eq:optgoal1}
    \min_{\bW} \sum_{i=1}^n (f_{\bW,\ba}(\bx_i)-y_i)^2 + \mu\|\bW\|_2^2.
\end{align}
We consider the GD training of \Eqref{eq:optgoal1}. Let
\begin{align*}
    \Phi(\bW) =\sum_{i=1}^n (f_{\bW,\ba}(\bx_i)-y_i)^2 + \mu\|\bW\|_2^2
\end{align*}
be the objective function in \Eqref{eq:optgoal1}. The gradient of $\Phi$ with respect to $\bw_r$ can be written as \citep{arora2019fine}
\begin{align*}
    \frac{\partial \Phi(\bW)}{\partial \bw_r} = \frac{2}{\sqrt{m}}a_r\sum_{i=1}^n (u_i - y_i)\mathbb{I}_{r,i}\bx_i + 2\mu\bw_r,\quad r\in[m],
\end{align*}
where $u_i=f_{\bW,\ba}(\bx_i)$ and $\mathbb{I}_{r,i} = \mathbb{I}\{\bw_r^\top\bx_i\geq 0\}$. Then the GD update rule is 
\begin{align*}
    \bw_r(k+1)=\bw_r(k)-\zeta\frac{\partial \Phi(\bW)}{\partial \bw_r}\biggm|_{\bW=\bW(k)},
\end{align*}
where $\bW(k)$ is the weight matrix at iteration $k$, and $\zeta$ is the learning rate. Define $\mathbb{I}_{r,i}(k)=\mathbb{I}\{\bw_r(k)^{\top}\bx_i\ge 0\}$, $\bZ(k)\in\RR^{md\times n}$ as
\begin{align*}
    \bZ(k)=\frac{1}{\sqrt{m}}
    \begin{pmatrix}
    a_1\mathbb{I}_{1,1}(k)\bx_1 & \dots & a_1\mathbb{I}_{1,n}(k)\bx_n\\
    \vdots & dots & \vdots \\
    a_m\mathbb{I}_{m,1}(k)\bx_1 & \dots & a_m\mathbb{I}_{m,n}(k)\bx_n
    \end{pmatrix},
\end{align*}
$\bH(k)=\bZ(k)^{\top}\bZ(k)$, and $\bu(k)=(\bu_1(k),...,\bu_n(k))^\top$ with $\bu_i(k)=f_{\bW(k),\ba}(\bx_i)$. Then the GD update rule with respect to $\bW$ can be written as
\begin{align}\label{modifiedgd}
    {\rm vec}(\bW(k+1)) =& {\rm vec}(\bW(k)) - 2\zeta\big( \bZ(k)(\bu(k)-\by) + \mu{\rm vec}(\bW(k))\big),
\end{align}
where $\mathrm{vec}(\bW) =(\bw_1^\top,\cdots,\bw_m^{\top})^\top\in \mathbb{R}^{md\times 1}$ is the vectorized weight matrix and $\by = (y_1,...,y_n)^\top$.

\paragraph{Neural Tangent Kernel (NTK)} It has been shown that the following NTK
\begin{align}\label{ntkh}
    h(\bs,\bt) = & \mathbb{E}_{\bw \sim N(0,\bI_d)}\rbr{\bs^\top\bt \ \mathbb{I}\{\bw^\top\bs \geq 0,\bw^\top\bt \geq 0\}}
    = \frac{\bs^\top\bt(\pi - \arccos(\bs^\top\bt))}{2\pi}
\end{align}
plays an important role in the study of one-hidden-layer ReLU neural networks, where $\bs,\bt$ are $d$-dimensional vectors \citep{du2018gradient,hu2021regularization}. Since $h$ is positive definite on the unit sphere $\mathbb{S}^{d-1}$ \citep{bietti2019inductive}, by Mercer's theorem, it possesses a Mercer decomposition as
$
    h(\bs,\bt) = \sum_{j=0}^\infty \lambda_j\varphi_j(\bs)\varphi_j(\bt),
$
where $\lambda_1\geq \lambda_2\geq...\geq 0$ are the eigenvalues, and $\{\varphi_j\}_{j=1}^{\infty}$ is an orthonormal basis. The asymptotic behavior of the eigenvalues is described in the following lemma.  
\begin{lemma}[Lemma 3.1 of \citet{hu2021regularization}]\label{lemeigendecay}
Let $\lambda_j$ be the eigenvalues of NTK $h$ defined above. Then we have $\lambda_j \asymp j^{-\frac{d}{d-1}}$.
\end{lemma} 
Let $\mathcal{N}$ denote the reproducing kernel Hilbert space (RKHS) generated by $h$ on $\mathbb{S}^{d-1}$, equipped with norm $\|\cdot\|_{\mathcal{N}}$. As a corollary of Lemma \ref{lemeigendecay}, it can be shown that the ($L_\infty$) entropy number of a unit ball in $\mathcal{N}$, denoted by $\mathcal{N}(1)$, can be controlled. The relationship between the eigenvalues and entropy numbers has been well studied; see \cite{edmunds2008function}.
\begin{lemma}\label{lementropy}
The entropy number of $\mathcal{N}(1)$, denoted by $H(\mathcal{N}(1),\delta,\|\cdot\|_{L_\infty})$, is bounded by $H(\mathcal{N}(1),\delta,\|\cdot\|_{L_\infty})\lesssim \delta^{-2(d-1)/d}$.
\end{lemma}

There are extensive works studying the generalization error bounds under NTK regime. For regression,
\citet{nitanda2020optimal,hu2021regularization} show the optimal convergence rates when using overparametrized one-hidden-layer neural networks, where the square loss is used. \citet{arora2019fine} provides generalization error bounds and provable learning scenarios for noiseless data. In the NTK regime, the neural network as a regressor is linked with the nonparametric regression via NTK. There are also other works studying the generalization performance of the neural network as a nonparametric regressor, out of the NTK regime; see \cite{schmidt2020nonparametric,farrell2021deep}. 
For classification, most of the existing results are established based on the separable data; see \cite{ji2019polylogarithmic,cao2019generalization,nitanda2019gradient} and references therein. In particular, \citet{hu2020simple} consider classification with noisy labels (labels are randomly flipped) and propose to use the square loss with $\ell_2$ regularization. 
Besides the generalization error bounds, another important research direction is to bridge the gap between NTK and finite-width overparametrized neural networks via GD training; see \cite{du2018gradient, arora2019fine, li2018learning,hu2021regularization}, among others.

\section{Overview of Reproducing Kernel Hilbert Space}
We provide here a brief overview of reproducing kernel Hilbert space (RKHS). 

\begin{definition}[Positive Definite Kernel]
A function $k:\Omega \times \Omega \mapsto \RR$ is said to be a \emph{positive definite kernel}, if $k(\bx, \tilde\bx) = k(\tilde\bx, \bx)$ for all $\bx,\tilde\bx\in\Omega$, and 
\[
\sum_{i=1}^n\sum_{j=1}^n \beta_i\beta_j k(\bx_i,\bx_j) > 0,
\]
for all $n\in\mathbb{N}$, $\beta_1,\ldots,\beta_n\in\RR$ such that at least one $\beta_j\neq 0$, and $\bx_1,\ldots,\bx_n\in\Omega$. 
\end{definition}

For a positive definite kernel $k$, define a linear space
\[
\mathcal{N}^0_k := \left\{\sum_{i=1}^n\beta_i k(\cdot,\bx_i):n\in\mathbb{N},\; \beta_1,\ldots,\beta_n\in\RR,\; \bx_1,\ldots,\bx_n\in\Omega\right\},
\]
and equip this space with an inner product $\langle \cdot, \cdot \rangle_{\mathcal{N}^0_k}$ by
\begin{eqnarray*}
\left\langle \sum_{i=1}^n\beta_i k(\cdot,\bx_i), \sum_{j=1}^{\tilde{n}}\tilde{\beta}_j k(\cdot, \tilde{\bx}_i)\right\rangle_{\mathcal{N}^0_k} := \sum_{i=1}^n\sum_{j=1}^{\tilde{n}}\beta_i\tilde{\beta}_j k(\bx_i, \tilde{\bx}_j).
\end{eqnarray*}
The norm of $g\in \mathcal{N}^0_k$ is defined by
$\|g\|_{\mathcal{N}^0_k}^2 := \langle g, g\rangle_{\mathcal{N}^0_k}$. Then the RKHS induced by $k$, denoted by $\mathcal{N}_k(\Omega)$, is defined as the closure of $\mathcal{N}^0_k(\Omega)$ with respect to the norm $\|\cdot\|_{\mathcal{N}^0_k(\Omega)}$.

For a subset $\Omega_0\subset \Omega$, define the \emph{restriction} of $\mathcal{N}_k$ on $\Omega_0$ as  
\[\mathcal{N}_k(\Omega_0) := \left\{g:\Omega_0\mapsto \RR: g = h|_{\Omega_0} \mbox{ for some }h\in\mathcal{N}_k \right\},\]
where $g=h|_{\Omega_0}$ means $g(\bx) = h(\bx)$ for all $\bx\in\Omega_0$. 
We equip $\mathcal{N}_k(\Omega_0)$ with norm  
\[\|g\|_{\mathcal{N}_k(\Omega_0)} := \inf_{\{h\in\mathcal{N}_k: h|_{\Omega_0} = g\}} \|h\|_{\mathcal{N}_k}. \]
Then, $\mathcal{N}_k(\Omega_0)$ is a RKHS with norm $\|\cdot\|_{\mathcal{N}_k(\Omega_0)}$ (see \citealt[page~351]{Aronszajn50}).

\section{Simplex Coordinates}
In simplex label coding, the one-hot labels are replaced by the simplex vertices of a $(K-1)$-simplex. The vertices of a regular $(K-1)$-simplex centered on the origin can be written as:
$$\bv_0 = \frac{1}{{\sqrt {2K}}}\cdot (1,\dots ,1)$$
 and for $1\leq i\leq K-1$, 
\begin{align*}
\bv_i=\frac{1}{\sqrt{2}}\be_{i}-\frac{1}{(K-1){\sqrt {2}}}\left(1 + \frac{1}{\sqrt{K}}\right)\cdot (1,\dots ,1). 
\end{align*}
The pairwise angle between vertices is $\arccos(-1/(K-1))$ and as $K\to\infty$, the angle converges to $90^{\circ}$. 

The vertices of a $(K-1)$-simplex can be viewed as maximally separated $K$ points on a sphere. 
In theory, the radius of the sphere doesn't matter but in practice, we recommend scaling it for larger number of classes, e.g., radius = $K$ for $K$-class classification. We find that such scaling empirically outperforms the default radius 1 in our experiments. More details can be found in Appendix \ref{app:realdata}.

\section{Assumptions}\label{app:assumptions}

In this work, we impose the following assumptions. In the rest of the Appendix, we use ${\rm poly}(t_1,t_2,\ldots)$ to denote some polynomial function with arguments $t_1,t_2,\ldots$. 

\begin{assumption}\label{as1}
Let $\lambda_{\min}(\bH^\infty)$ be the minimum eigenvalue of the symmetric matrix $\bH^\infty$, where $\bH^{\infty}=\rbr{h(\bx_i,\bx_j)}_{n\times n}$ ($\bH^{\infty}$ is usually called the NTK matrix). 
Let $\lambda_0$ be the largest number such that with probability at least $1-\delta_n$, $\lambda_{\min}(\bH^\infty)\geq \lambda_0$, and $\delta_n \rightarrow 0$ as $n$ goes to infinity\footnote{Potential dependency of $\lambda_0$ on $n$ is suppressed for notational simplicity.}. For sufficiently large $n$, the regularization parameter $\mu \asymp n^{\frac{d-1}{2d-1}}$, the learning rate $\zeta = o(n^{-\frac{3d-1}{2d-1}})$, the variance of initialization $\xi^2=O(1)$, the number of nodes in the hidden layer $m\geq \xi^{-2}{\rm ploy}(n,\lambda_0^{-1})$, and the iteration number $k$ satisfies 
$\log \rbr{{\rm ploy}_1(n,\xi,1/\lambda_0)} \lesssim \zeta\mu k \lesssim \log \rbr{{\rm ploy}_2(\xi,1/n,\sqrt{m})}.$ 
\end{assumption}

\begin{assumption}\label{as2}
The conditional probability in the non-separable case satisfies $\eta\in \mathcal{N}$.
\end{assumption}

\begin{assumption}\label{as4}
The solution to \Eqref{eq:optgoal} satisfies $\|f_{\bW(k),\ba}\|_{\mathcal{N}}\leq C$, where $C$ is a constant not depending on $n$.
\end{assumption}

\begin{remark}
Assumption \ref{as4} can be replaced by a stronger assumption, that is, $f_{\bW(k),\ba}$ has a bounded Lipschitz constant, and the constant does not depend on $n$.
\end{remark}

\begin{assumption}\label{as6}
The probability density function of the marginal distribution $P_X$, denoted by $p(\vx)$, is continuous on $\Omega$, and there exists a positive constant $c_0$ such that
\begin{align*}
    p(\vx) \leq c_0, \forall \vx\in \Omega.
\end{align*}
\end{assumption}

\begin{assumption}\label{as5}
The probability density function of the marginal distribution $p(\vx)$ is continuous on $\Omega$, and there exist positive constants $c_1\leq c_2$ such that
\begin{align*}
    c_1 \leq p(\vx) \leq c_2, \forall \vx\in \Omega.
\end{align*}
\end{assumption}

Assumption \ref{as1} is related to the neural network and GD training, where similar settings have been adopted by \citet{arora2019fine,hu2021regularization}. From the results in \cite{arora2019fine,hu2021regularization}, the width of the neural network depends on the minimum eigenvalue of the NTK matrix $\lambda_{\min}(\bH^\infty)$, where a smaller $\lambda_{\min}(\bH^\infty)$ leads to a wider neural network. Therefore, it is desired that $\lambda_0$ is as large as possible. However, the consistency requires that the probability is tending to one; thus, we require $\delta_n\rightarrow 0$ as $n\rightarrow \infty$. As $n$ becomes larger, with probability tending to one, the distance of the two nearest points in $n$ input points converges to zero, thus making $\bH^\infty$ close to a degenerate matrix, and the minimum eigenvalue of $\bH^\infty$ converges to zero. Therefore, inevitably, $\lambda_0\rightarrow 0$ (but $\lambda_{\min}(\bH^\infty)$ is strictly larger than 0 for all $n$ with probability one). The requirements of the regularization parameter, the learning rate, the variance of initialization, the number of nodes in the hidden layer and the iteration number are all the same as those in \cite{hu2021regularization}.

Assumption \ref{as2} imposes conditions on the underlying true conditional probability in the \textit{non-separable} case. This assumption basically requires that the conditional probability is within the function class generated by the GD-trained neural networks we consider (thus can be calibrated). Given that the neural networks are highly flexible, we believe that most of the functions are within the function class generated by the neural networks. As a simple example, any Lipschitz functions are within this function class.

Assumption \ref{as4} requires that the solution to \Eqref{eq:optgoal} is well-behaved, i.e., the solution is within a ball in $\mathcal{N}$ with a certain radius. Roughly speaking, Assumption \ref{as4} requires that the \textit{complexity} of the neural network estimator generated by the GD training is controlled. Since the step size is relatively small and the iteration number is not large (only $\log({\rm poly}(n,\xi, ,\lambda_0^{-1})$), we believe it is a mild assumption.

Assumption \ref{as6} only requires the probability to be upper bounded from infinity, while Assumption \ref{as5} requires the probability to be upper bounded from infinity and lower bounded away from zero on the support $\Omega$.  They are standard assumptions used in the classical analysis of classification in statistics; see \citet{audibert2007fast,kohler2007rate} for example. Clearly, uniform distribution satisfies Assumptions \ref{as6} and \ref{as5}. In \citet{audibert2007fast}, Assumption \ref{as6} is called mild density assumption and Assumption \ref{as5} is called strong density assumption.

\section{Proofs of Main Results}
\label{sec:appendix-proof}

This section includes the proofs of main results in the paper.

\subsection{Proof of Theorem \ref{thm:nonsepthm1}}

We first introduce some lemmas that are used in the proof of Theorem \ref{thm:nonsepthm1}.

Let $l_1(y_i,f(\bx_i)) = (1-y_if(\bx_i))^2 =(y_i-f(\bx_i))^2$ be the square loss on a training point $(\bx_i,y_i)$, the $l_1$-risk of $f$ be $\mathcal{R}_{l_1}(f) = \mathbb{E}_{X,Y\sim P}l_1(Y,f(X))$, and $\mathcal{R}_{l_1} = \min_{f\in \mathcal{N}}\mathbb{E}_{X,Y\sim P}l_1(Y,f(X))$. Let $L_1(f, \bx,y) = \mu\|f\|_{\mathcal{N}}^2 + l_1(y,f(\bx))$ and the $L_1$-risk of $f$ be $ \mathcal{R}_{L_1}(f) = \mathbb{E}_{X,Y\sim P}L_{1}(f,X,Y)$. Let $f_n =\argmin_{f\in \mathcal{N}}\mathcal{R}_{L_1}(f)$. 

Lemma \ref{lem:stein} is (a weaker version of) Theorem 5.6 of \citet{steinwart2007fast}, which provides a bound on the deviation between the empirical minimizer and true minimizer. Lemma \ref{lem:condofstein} is used to verify that one of the conditions of Lemma \ref{lem:stein} is fulfilled. Lemma \ref{thm:withPenalty} shows that under certain conditions, the solution to 
\begin{align}\label{regpro}
    \min_{f\in \cN}\frac{1}{n}\sum_{i=1}^n(y_i - f(\bx_i))^2 + \frac{\mu}{n}\|f\|_{\mathcal{N}}^2
\end{align}
is closely related to the estimator given by the overparameterized neural networks $f_{\bW(k),\ba}$. Lemma \ref{thm:withPenalty} can be obtained by merely repeating the proof of Theorem 5.2 of \citet{hu2021regularization}, since we require that the probability density function $p(\vx)$ of $P_X$ is upper bounded by a positive constant by Assumption \ref{as6}. Therefore, the only difference is that we replace $\|\cdot\|_{2}$ (which corresponds to the uniform distribution) to the $L_2$ norm corresponding to the probability measure $P_X$; thus the proof is omitted. Note also that the second statement of Lemma \ref{thm:withPenalty} corresponds to the noiseless case.

\begin{lemma}\label{lem:stein} 
Let $Z=\Omega\times \{-1,1\}$. Let $\mathcal{F}$ be a convex set of bounded measurable functions from $Z$ to $\mathbb{R}$ and let $L:\mathcal{F}\times Z \rightarrow [0,\infty)$ be a convex and continuous loss function. For a probability measure $P$ on $Z$, define 
$$\mathcal{G}:=\{L \circ f - L \circ {f}_{P,\mathcal{F}} : f \in \mathcal{F}\},$$
where ${f}_{P,\mathcal{F}}$ is a minimizer of $\mathbb{E}_{Z\sim P}L(f,Z)$.
Suppose that there are constants $c \ge 0$, $0 < \alpha \le 1$, $\delta \le 0$ and $B >0 $ such that $\mathbb{E}_{Z\sim P}{g}^{2} \le c(\mathbb{E}_{Z\sim P}g)^{\alpha} + \delta$ and $\vert|g\vert|_{\infty} \le B$ for all  $g \in \mathcal{G}$. Furthermore, assume that $\mathcal{G}$ is separable with respect to $\|\cdot\|_{\infty}$ and that there are constants $a\geq 1$ and $0<\alpha<2$ with
\begin{align}\label{entropyblem}
    \sup_{T\in Z^n} H(B^{-1}\mathcal{G},\epsilon,\|\cdot\|_{L_2(T)})\leq a\epsilon^{-\beta}
\end{align}
for all $\epsilon>0$, where $H(B^{-1}\mathcal{G},\epsilon,\|\cdot\|_{L_2(T)})$ is the entropy number of the set $B^{-1}\mathcal{G}$, and $\|f\|_{L_2(T)}^2=\frac{1}{n}\sum_{i=1}^n f(\vx_i,y_i)^2$ is the empirical norm. Then there exists a constant $c_{\beta} > 0$ depending only on $\beta$ such that for all $n\ge 1$ and all $t \ge 1$ we have
\begin{align*}
    \PP(T\in Z^n:\mathcal{R}_{L,P}(f_{T,\mathcal{F}}) > \mathcal{R}_{L,P}(f_{P,\mathcal{F}}) + c_\beta \varepsilon(n,a,B,c,\delta,t))\leq e^{-t},
\end{align*}
where 
\begin{align}\label{lemsteps}
& \varepsilon(n, a, B, c, \delta, t) \nonumber\\
= & B^{2 \beta /(4-2 \alpha+\alpha p)} c^{(2-\beta) /(4-2 \alpha+\alpha \beta)}\left(\frac{a}{n}\right)^{2 /(4-2 \alpha+\alpha \beta)}+B^{\beta / 2} \delta^{(2-\beta) / 4}\left(\frac{a}{n}\right)^{1/2}\nonumber\\
& + B\left(\frac{a}{n}\right)^{2/(2+\beta)} + \sqrt{\frac{\delta t}{n}} + \left(\frac{ct}{n}\right)^{1/(2-\alpha)} + \frac{Bt}{n},
\end{align}
and $f_{T,\mathcal{F}}$ is the minimizer with respect to the empirical measure. 
\end{lemma}

\begin{lemma}\label{lem:condofstein}
Assume the conditions of Theorem \ref{thm:nonsepthm1} hold. Define $C := 8\vert|(2\eta - 1)^{-1} \vert|_{\kappa,\infty} + 32$, where $\|\cdot\|_{\kappa,\infty}$ is the norm of Lorentz space $L_{\kappa,\infty}$ \citep{bennett1988interpolation}. Let $\mu > 0$ and $0 < \gamma \le n^{1/2}\mu^{-1/2}$, then for all $f \in \gamma \mathcal{N}(1)$ we have
\begin{equation*}
     \mathbb{E}_{X,Y\sim P}(L_{1} \circ f - L_{1} \circ f_{n})^2 \leq C(K\gamma + 1)^2(\mathbb{E}_{X,Y\sim P}(L_{1} \circ f - L_{1} \circ f_{n})) + 2C(K\gamma + 1)^{2}a(\mu),
\end{equation*}
where $a(\mu)$ is the approximation error function given by 
\begin{equation*}
    a(\mu) = \inf_{f \in \mathcal{N}}(n^{-1}\mu \vert|f\vert|_{\mathcal{N}}^{2} + \mathcal{R}_{l_1}(f) - \mathcal{R}_{l_1}).
\end{equation*}
\end{lemma}

\begin{lemma}\label{thm:withPenalty}
Suppose Assumptions \ref{as1} and \ref{as6} hold. Then we have 
\[\mathbb{E}_{X\sim P_X}(f_{\bW(k),\ba}(X)-\hat f(X))^2 = O_{\mathbb{P}}(n^{-\frac{d}{2d-1}}),\] where $\hat f$ is the solution to \Eqref{regpro}. 
Furthermore, if there exists a function $f\in \mathcal{N}$ that does not depend on $n$ and $f(\vx_i)=y_i$ for all $i=1,...,n$, then we can set $\mu=o(1)$ and obtain \[\mathbb{E}_{X\sim P_X}(f_{\bW(k),\ba}(X)-\hat f(X))^2 = o_{\mathbb{P}}(1).\]  
\end{lemma}

\begin{remark}
According to the proof in \cite{hu2021regularization}, the probability in $o_{\mathbb{P}}(1)$ of Lemma \ref{thm:withPenalty} only relates to the width of the one-hidden-layer neural network, which can be arbitrarily small by enlarging the neural network's width.
\end{remark}

Now we are ready to prove Theorem \ref{thm:nonsepthm1}. Let $L=L_1$ in Lemma \ref{lem:stein}, which is clearly continuous. Let $\hat f$ be the solution to \Eqref{regpro}. The key idea in this proof is using $\hat f$ to bridge two functions $2\eta-1$ and $f_{\bW(k),\ba}$.

Since $\hat f$ is the solution to \Eqref{regpro}, it can be seen that
\begin{align}\label{eq:basicineq}
    \frac{1}{n}\sum_{i=1}^n(y_i - \hat f(\vx_i))^2 + \frac{\mu}{n}\|\hat f\|_{\mathcal{N}}^2 \leq & \frac{1}{n}\sum_{i=1}^n(y_i - (2\eta(\vx_i)-1))^2 + \frac{\mu}{n}\|2\eta(\vx_i)-1\|_{\mathcal{N}}^2\nonumber\\
    \leq & \frac{2}{n}\sum_{i=1}^n(y_i^2 + (2\eta(\vx_i)-1)^2) + \frac{\mu}{n}\|2\eta(\vx_i)-1\|_{\mathcal{N}}^2
    \nonumber\\
    \leq & C_1,
\end{align}
where the second inequality is by the Cauchy-Schwarz inequality, and the third inequality is because $y_i^2=1$ and $\eta(\vx)$ is bounded.

The reproducing property implies that
\begin{align*}
    \hat f(\vx) = \langle \hat f,h(\vx,\cdot) \rangle_{\mathcal{N}}\leq \|\hat f\|_{\mathcal{N}}\|h(\vx,\cdot)\|_{\mathcal{N}}=\|\hat f\|_{\mathcal{N}}\sqrt{h(\vx,\vx)}, \forall \vx\in \Omega,
\end{align*}
which yields
\begin{align*}
    \|\hat f\|_{L_\infty}\leq C_2\|\hat f\|_{\mathcal{N}}.
\end{align*}
Together with \Eqref{eq:basicineq}, we obtain 
\begin{align}\label{eq:Nbound}
    \|\hat f\|_{L_\infty}\leq C_3\|\hat f\|_{\mathcal{N}}\leq C_4(\mu/n)^{-1/2}.
\end{align}
Thus, we can take $B=C_4(\mu/n)^{-1/2}$ in Lemma \ref{lem:stein}. The entropy condition can be verified via Lemma \ref{lementropy}, which allows us to take $\beta = 2(d-1)/d$. \Eqref{eq:Nbound}, together with Lemma \ref{lem:condofstein}, also suggests that we can take $c=C(KB + 1)^2$, $\alpha = 1$, and $\delta = 2C(KB + 1)^{2}a(\mu)$. 

Next, we provide an upper bound on $a(\mu)$. The definition of $a(\mu)$ implies
\begin{align}\label{eq:sepamuub1}
    a(\mu) = & n^{-1}\mu\|f_n\|_{\mathcal{N}}^2 + R_{l_1}(f_n)-R_{l_1}\nonumber\\
    = & n^{-1}\mu\|f_n\|_{\mathcal{N}}^2 + \mathbb{E}_{X\sim P_X} (2\eta(X) - 1 - f_n(X))^2\nonumber\\
    \leq & n^{-1}\mu \|2\eta-1\|_{\mathcal{N}}^2, 
\end{align}
where we use the relationship $\mathcal{R}_{l_1,P}(f) - \mathcal{R}_{l_1,P}=\mathbb{E}_{X\sim P_X}(2\eta(X)-1-f(X))^2$.

Plugging all the terms into \Eqref{lemsteps}, together with Lemma \ref{lem:stein}, yields that 
\begin{align}\label{eq:sepub1}
    \mathcal{R}_{L_1,P}(\hat f) = \mathcal{R}_{L_1,P}(f_n) + O_{\mathbb{P}}( \varepsilon(n,a,B,c,\delta)),
\end{align}
where 
\begin{align}\label{eq:epsub1}
    \varepsilon(n, a, B, c, \delta) = &  B^{\frac{4}{2+\beta}}n^{-\frac{2}{2+\beta}} + B (\mu/n)^{\frac{2-\beta}{4}}n^{-\frac{1}{2}}\|2\eta-1\|_{\mathcal{N}}^{\frac{2-\beta}{2}}+ B^{2}n^{-1}\nonumber\\
    = & B^{\frac{4d}{4d-2}}n^{-\frac{2d}{4d-2}} + B \mu^{\frac{1}{2d}}n^{-\frac{1}{2}-\frac{1}{2d}}\|2\eta-1\|_{\mathcal{N}}^{\frac{1}{d}}+ B^{2}n^{-1}.
\end{align}

Since $\mathcal{R}_{l_1,P}(f) - \mathcal{R}_{l_1,P}=\mathbb{E}_{X\sim P_X}(2\eta(X)-1-f(X))^2$, we subtract $\mathcal{R}_{l_1,P}$ on both sides of \Eqref{eq:sepub1} and get
\begin{align}\label{eq:sepub2}
    & \mathbb{E}_{X\sim P_X}(2\eta(X)-1-\hat f(X))^2 + n^{-1}\mu\|\hat f\|_{\mathcal{N}}^2\nonumber\\
    = & \mathbb{E}_{X\sim P_X}(2\eta(X)-1-f_n(X))^2 + n^{-1}\mu\|f_n\|_{\mathcal{N}}^2 + O_{\mathbb{P}}( \varepsilon(n,a,B,c,\delta))\nonumber\\
    = & O_{\mathbb{P}}(n^{-1}\mu\|2\eta-1\|_{\mathcal{N}}^2 + \varepsilon(n,a,B,c,\delta)),
\end{align}
where the last equality (with big $O$ notation) is by \Eqref{eq:sepamuub1}. Combining \Eqref{eq:sepub2} and  \Eqref{eq:Nbound} implies
\begin{align*}
    \|\hat f\|_{L_\infty}^2 \leq C_3^2 \|\hat f\|_{\mathcal{N}}^2 = O_{\mathbb{P}}(1 + n\mu^{-1}\varepsilon(n,a,B,c,\delta)).
\end{align*}
In the following, we will show that by taking $\mu \asymp n^{\frac{d-1}{2d-1}}$,
\begin{align}\label{eq:sepub23}
    \mathbb{E}_{X\sim P_X}(2\eta(X)-1-\hat f(X))^2 + n^{-1}\mu\|\hat f\|_{\mathcal{N}}^2=O_{\mathbb{P}}(n^{-\frac{d}{2d-1}}\max(1,\|2\eta-1\|_{\mathcal{N}}^{\frac{2}{d}})) = O_{\mathbb{P}}(n^{-\frac{d}{2d-1}}).
\end{align}

If $n\mu^{-1}\varepsilon(n,a,B,c,\delta)\lesssim 1$, then $\varepsilon(n,a,B,c,\delta)\lesssim \mu/n$, and \Eqref{eq:sepub23} holds. Otherwise, we can replace $B^2$ by its upper bound $O_{\mathbb{P}}(n\mu^{-1}\varepsilon(n,a,B,c,\delta))$ in \Eqref{eq:epsub1} and obtain that
\begin{align*}
    \varepsilon = O_{\mathbb{P}}(\varepsilon^{\frac{d}{2d-1}} \mu^{-\frac{d}{2d-1}} + \varepsilon^{\frac{1}{2}}\mu^{-\frac{d-1}{2d}}n^{-\frac{1}{2d}}\|2\eta-1\|_{\mathcal{N}}^{\frac{1}{d}}),
\end{align*}
where we set $\varepsilon = \varepsilon(n,a,B,c,\delta)$ for notational simplicity. Let us hereby denote $I_{1} = \varepsilon^{\frac{d}{2d-1}} \mu^{-\frac{d}{2d-1}}$ and $I_{2} = \varepsilon^{\frac{1}{2}}\mu^{-\frac{d-1}{2d}}n^{-\frac{1}{2d}}\|2\eta-1\|_{\mathcal{N}}^{\frac{1}{d}}$, and consider the following two cases.

\textbf{Case 1:} $I_{1} \ge I_{2}$, 
then we have 
\begin{align*}
    \varepsilon=O_{\mathbb{P}}(\varepsilon^{\frac{d}{2d-1}} \mu^{-\frac{d}{2d-1}}).
\end{align*}
Solving this equality leads to 
\begin{align}\label{eq:epsC1ub1}
    \varepsilon=O_{\mathbb{P}}(\mu^{-\frac{d}{d-1}}).
\end{align}
Plugging \Eqref{eq:epsC1ub1} into \Eqref{eq:sepub2} and minimize the right-hand side of \Eqref{eq:sepub2} with respect to $\mu$ gives us $\mu \asymp n^{\frac{d-1}{2d-1}}$; thus \Eqref{eq:sepub23} holds.

\textbf{Case 2:} $I_{1} < I_{2}$, then we have 
\begin{align*}
    \varepsilon=O_{\mathbb{P}}(\varepsilon^{\frac{1}{2}}\mu^{-\frac{d-1}{2d}}n^{-\frac{1}{2d}}\|2\eta-1\|_{\mathcal{N}}^{\frac{1}{d}}),
\end{align*}
which leads to
\begin{align}\label{eq:epsC2ub1}
    \varepsilon=O_{\mathbb{P}}(\mu^{-\frac{d-1}{d}}n^{-\frac{1}{d}}\|2\eta-1\|_{\mathcal{N}}^{\frac{2}{d}}).
\end{align}
Similarly, we plug \Eqref{eq:epsC2ub1} into \Eqref{eq:sepub2} and minimize the right-hand side of \Eqref{eq:sepub2} with respect to $\mu$ and obtain $\mu \asymp n^{\frac{d-1}{2d-1}}$, which also leads to \Eqref{eq:sepub23}.

Now we can obtain an upper bound on the excess risk. For the notation simplicity, let $f=f_{\bW(k),\ba}$. The excess risk can be bounded by
\begin{align}\label{eq:misratesep1}
     L(f) - L^{*}  \leq &  
    \mathbb{E}_{X\sim P_X}\mathbb{I}\{(2\eta(X) - 1)f(X) \leq 0, |\eta(X) -0.5|<\delta\}|2\eta(X) - 1| \nonumber\\
    & + \mathbb{E}_{X\sim P_X}\mathbb{I}\{(2\eta(X) - 1)f(X) \leq 0, |\eta(X) -0.5|\ge \delta\}|2\eta(X) - 1|.
\end{align}
The first term can be bounded via Tsybakov's noise condition as
\begin{align}\label{eq:misratesep1I1}
    & \mathbb{E}_{X\sim P_X}\mathbb{I}\{(2\eta(X) - 1)f(X) \leq 0, |\eta(X) -0.5|<\delta\}|2\eta(X) - 1|  \le 2\delta \mathbb{E}[\mathbb{I}\{|\eta(X) - 0.5|<\delta\}]\nonumber\\
    = & 2\delta\mathbb{P}(|\eta(X) - 0.5|<\delta) \leq 2C\delta^{\kappa+1}.
\end{align}

It remains to bound the second term in \Eqref{eq:misratesep1}. If $p(\bx)$ is continuous, then by the fact that $|2\eta(X) - 1|\leq |2\eta(X) - 1-f(X)|$ if $(2\eta(X) - 1)f(X) \leq 0$, we have
\begin{align}\label{eq:misratesep1I2}
    & \mathbb{E}_{X\sim P_X}\mathbb{I}\{(2\eta(X) - 1)f(X) \leq 0, |\eta(X) -0.5|\ge \delta\}|2\eta(X) - 1|\nonumber\\
    \leq & 2\delta^{-1}\mathbb{E}_{X\sim P_X}\mathbb{I}\{(2\eta(X) - 1)f(X) \leq 0, |\eta(X) -0.5|\ge \delta\}|2\eta(X) - 1|^2 \nonumber\\
    \leq & 2\delta^{-1}\mathbb{E}_{X\sim P_X}\mathbb{I}\{|\eta(X) -0.5|\ge \delta\}(2\eta(X) - 1-f(X))^2\nonumber\\
    \leq & 2\delta^{-1}\mathbb{E}_{X\sim P_X}(2\eta(X) - 1-f(X))^2\nonumber\\
    \leq & 4 \delta^{-1}\mathbb{E}_{X\sim P_X}(2\eta(X) - 1-\hat f(X))^2 + 4 \delta^{-1}\mathbb{E}_{X\sim P_X}(f(X)-\hat f(X))^2\nonumber\\
    = & O_{\mathbb{P}}(\delta^{-1}n^{-\frac{d}{2d-1}}),
\end{align}
where the fourth inequality is by the Cauchy-Schwarz inequality, and the last equality (with big $O$ notation) is by \Eqref{eq:sepub23} and Lemma \ref{thm:withPenalty}. 
Taking $\delta = n^{-\frac{d}{(2d-1)(\kappa + 2)}}$, and plugging \Eqref{eq:misratesep1I1} and \Eqref{eq:misratesep1I2} into \Eqref{eq:misratesep1} leads to 
\begin{align*}
    L(f) = L^{\star} + O_{\mathbb{P}}(n^{-\frac{d(\kappa+1)}{(2d-1)(\kappa + 2)}}).
\end{align*}
This finishes the proof.

\subsection{Proof of Theorem \ref{thm:sep}}
We first present a lemma.
\begin{lemma}\label{lem:as3}
Suppose two sets are separable with a positive margin $\gamma>0$. Then there exists a function $f_T$ satisfying
\begin{align*}
    f_T(\vx)=1, \forall \vx\in \Omega_1, \quad f_T(\vx)=-1, \forall \vx\in \Omega_2.
\end{align*}
\end{lemma}

\textit{Proof of Theorem \ref{thm:sep}.}
By the equivalence of the RKHS generated by the Laplace kernel and $\mathcal{N}$ \citep{geifman2020similarity,chen2020deep}, it can be shown that $\mathcal{N}$ can be embedded into the Sobolev space $W_2^{\nu}$ for some $\nu>d/2$. Consider the H\"older space $C^{0,\alpha}_b$ for $0< \alpha\leq 1$ equipped with the norm
\begin{align}\label{eq:holdnorm}
    \|f\|_{C_b^{0,\alpha}}:=\sup_{\vx,\vx'\in\Omega,\vx\neq \vx'}\frac{|f(\vx)-f(\vx')|}{\|\vx-\vx'\|_2^\alpha}. 
\end{align}
By the Sobolev embedding theorem, we have the embedding relationship
\begin{align}\label{eq:embedding}
\|f\|_{C_b^{0,\tau}}\leq C_1\|f\|_{W^{\nu}_2}\leq C_2 \|f\|_{\mathcal{N}}
\end{align}
for all $f\in \mathcal{N}$, where $\tau = \min(\nu-d/2,1)$. 

Without loss of generality, let us consider $\vx\in \Omega_1$. The case of $\vx\in \Omega_2$ can be proved similarly. For any $\vx\in \Omega_1$, take $\vx'=\argmin_{\vx_i} \|\vx_i-\vx\|_2$. Thus, the definition of the H\"older space and \Eqref{eq:embedding} imply
\begin{align}\label{eq:lbsep2}
    |f_{\bW(k),\ba}(\vx) - f_{\bW(k),\ba}(\vx')| \leq C_2\|f_{\bW(k),\ba}\|_{\mathcal{N}} \|\vx'-\vx\|_2^\tau \leq C_3\|\vx'-\vx\|_2^\tau,
\end{align}
where the last inequality is by Assumption \ref{as4}.

Let $\hat f$ be the solution to \Eqref{regpro}, and let $\hat f_1$ be the solution to \Eqref{regpro} with $\mu=0$. Let $f_T$ be as in Lemma \ref{lem:as3}. Note that $\hat f_1$ satisfies $\hat f_1(\vx_i)=f_T(\vx_i)$. Thus, by the identity $\|\hat f_1\|_{\mathcal{N}}^2 + \|\hat f_1-f_T\|_{\mathcal{N}}^2 = \|\hat f_T\|_{\mathcal{N}}^2$ \citep{wendland2004scattered}, 
we have $\|\hat f_1\|_{\mathcal{N}}\leq \|f_T\|_{\mathcal{N}}$.
Since $\hat f$ is the solution to \Eqref{regpro}, we have \begin{align}\label{eq:basicineqsep}
    \frac{1}{n}\sum_{i=1}^n(y_i - \hat f(\vx_i))^2 + \frac{\mu}{n}\|\hat f\|_{\mathcal{N}}^2 \leq & \frac{1}{n}\sum_{i=1}^n(y_i - f_T(\vx_i))^2 + \frac{\mu}{n}\|f_T\|_{\mathcal{N}}^2=\frac{\mu}{n}\|f_T\|_{\mathcal{N}}^2,
\end{align}
which implies $\|\hat f\|_{\mathcal{N}}\leq \|f_T\|_{\mathcal{N}}$, where we utilize $y_i = f_T(\vx_i)$ in the separable case.

Direct computation shows
\begin{align}\label{eq:lbsep1}
    f_{\bW(k),\ba}(\vx') = & \hat f_1(\vx') -(\hat f_1(\vx')- \hat f(\vx')) - (\hat f(\vx')-f_{\bW(k),\ba}(\vx'))\nonumber\\
    = & 1- I_1 - I_2,
\end{align}
where we use $\hat f_1(\vx_i)=1$ for any $\vx_i\in \Omega$; thus $\hat f_1(\vx')=1$.

By the representer theorem, $\hat f$ and $\hat f_1$ can be expressed as
\begin{align*}
    \hat f_1(\vx) = h(\bx,\bX)(\bH^\infty + \mu \bI_n)^{-1}\by, \hat f(\vx) = h(\bx,\bX)(\bH^\infty)^{-1}\by,
\end{align*}
where 
$h(\bx,\bX) = (h(\bx,\bx_1),...,h(\bx,\bx_n))\in\RR^{1\times n}$,  $\bH^{\infty}=\rbr{h(\bx_i,\bx_j)}_{n\times n}$, and $\by=(y_1,...,y_n)^\top = (f_T(\vx_1),...,f_T(\vx_n))^\top$. 
Thus, the first term $I_1$ in \Eqref{eq:lbsep1} can be bounded by
\begin{align}\label{eq:lbsep122}
   |I_1| = & |\hat f_1(\vx') - \hat f(\vx')| = |h(\bx',\bX)(\bH^\infty)^{-1}\by - h(\bx',\bX)(\bH^\infty + \mu \bI_n)^{-1}\by|\nonumber\\
   = & |\mu h(\bx',\bX)(\bH^\infty)^{-1}(\bH^\infty + \mu \bI_n)^{-1}\by|\nonumber\\
   \leq & \mu \sqrt{h(\bx',\bX)(\bH^\infty)^{-1}(\bH^\infty + \mu \bI_n)^{-1}(\bH^\infty)^{-1}h(\bx',\bX)^\top\by^\top (\bH^\infty + \mu \bI_n)^{-1} \by}\nonumber\\
   = & \mu \sqrt{h(\bx',\bX)(\bH^\infty)^{-1}(\bH^\infty + \mu \bI_n)^{-1}(\bH^\infty)^{-1}h(\bx',\bX)^\top}\|\hat f_1\|_{\mathcal{N}}\nonumber\\
   \leq & \sqrt{\mu} \sqrt{h(\bx',\bX)(\bH^\infty)^{-2}h(\bx',\bX)^\top}\|f_T\|_{\mathcal{N}} = \sqrt{\mu}\|f_T\|_{\mathcal{N}},
\end{align}
where the first inequality is by the Cauchy-Schwarz inequality, the second inequality is because $(\bH^\infty + \mu \bI_n)^{-1} \preceq \mu^{-1}\bI_n$, and the last equality is because for any $\vx_i$, $(\bH^\infty)^{-1}h(\bx_i,\bX)^\top=\be_i$. Therefore, $I_1$ converges to zero as $n\rightarrow \infty$ since $\mu=o(1)$. Specifically, there exists an $n_1$ such that when $n\geq n_1$, $|I_1|\leq 1/4$.

The second term $I_2$ in \Eqref{eq:lbsep1} can be bounded by
\begin{align}\label{eq:lbsepI2}
    |I_2|\leq & \|\hat f-f_{\bW(k),\ba}\|_\infty \leq C_4\|\hat f-f_{\bW(k),\ba}\|_{\mathcal{N}}^{\frac{d-1}{d}}\|\hat f-f_{\bW(k),\ba}\|_{2}^{\frac{1}{d}}\nonumber\\
    \leq & C_4(\|\hat f\|_{\mathcal{N}}+\|f_{\bW(k),\ba}\|_{\mathcal{N}})^{\frac{d-1}{d}}\|\hat f-f_{\bW(k),\ba}\|_{2}^{\frac{1}{d}}\nonumber\\
    \leq & C_5\left(\mathbb{E}_{X\sim P_X}(\hat f(X)-f_{\bW(k),\ba}(X))^2\right)^{\frac{1}{2d}},
\end{align}
which converges to zero by Lemma \ref{thm:withPenalty}. In \Eqref{eq:lbsepI2}, the second inequality is by the interpolation inequality, the third inequality is by the triangle inequality, and the last inequality is because of Assumption \ref{as5}. Therefore, there exists an $n_2$ such that when $n\geq n_2$, with probability at least $1-\delta$, $|I_2|\leq 1/4$.

Take $n_0=\max(n_1,n_2)$. For $n\geq n_0$, \Eqref{eq:lbsep1} gives us $ f_{\bW(k),\ba}(\vx')\geq 1/2$ with probability at least $1-\delta$. Therefore, by \Eqref{eq:lbsep2}, as long as 
\begin{align}\label{eq:ubsep1}
   \|\vx'-\vx\|_2 = \min_{\bx_i}\|\vx_i-\vx\|_2 \leq (4C_3)^{-1/\tau} :=C_6, \forall \vx\in\Omega_1,
\end{align}
we have $f_{\bW(k),\ba}(\vx)\geq 1/4$ for all $\bx\in \Omega_1$, which implies that the missclassification rate is zero. 

Let ${\rm N}(\delta,\Omega_{1},\|\cdot\|_2)$ be the covering number of $\Omega_{1}$ and ${\rm N}_0 = {\rm N}(C_6/2,\Omega_1,\|\cdot\|_2)$. Since $\Omega_1$ is compact and $C_6>0$, ${\rm N}_0$ is finite (and is a constant). Therefore, $\Omega_1$ can be covered by ${\rm N}_0$ balls with radius $C_6/2$ (denoted by $\mathbf{B}$), and as long as for each ball, there exists one point $\vx_j$ in this ball, \Eqref{eq:ubsep1} is satisfied. Since $\vx_k$ has a probability density function with lower bound $c_1$, it remains to upper bound the probability that there exists one ball such that there is no point in it. Define this event as $\mathcal{A}$. The union bound of probability implies that for $n>n_0$,
\begin{align*}
    \mathbb{P}(\mathcal{A}) \leq {\rm N}_0 \left(1-\frac{c_1Vol(\mathbf{B})}{Vol(\Omega_1)}\right)^n \leq {\rm N}_0 \exp(-C_7n),
\end{align*}
where $C_7 = - \log \left(1-\frac{c_1Vol(\mathbf{B})}{Vol(\Omega_1)}\right)$ is a positive constant. Clearly, we can adjust the constants such that the results in Theorem \ref{thm:sep} holds for all $n$. This finishes the proof.

\subsection{Proof of Theorem \ref{prop:sep}}
Note that $f_{\bW(k),\ba}$ is a classifier, and the decision boundary is defined by $\mathcal{D}_T:=\{\vx|f_{\bW(k),\ba}(\vx)=0\}$. Take any point $\vx'$ in $\mathcal{D}_T$. The definition of the H\"older space and \Eqref{eq:embedding} imply
\begin{align}\label{eq:sepeq11}
    \frac{|f_{\bW(k),\ba}(\vx)-f_{\bW(k),\ba}(\vx')|}{\|\vx-\vx'\|_2^\tau}\leq C_2 \|f_{\bW(k),\ba}\|_{\mathcal{N}} \leq C_3, \forall \vx\in \Omega,
\end{align}
which is the same as
\begin{align}\label{eq:sepeq1}
    \|\vx-\vx'\|_2^\tau\geq |f_{\bW(k),\ba}(\vx)|/C_3, \forall \vx\in \Omega,
\end{align}
where the last inequality in \Eqref{eq:sepeq11} is because of Assumption \ref{as4}. Therefore, it suffices to provide a lower bound of $|f_{\bW(k),\ba}(\vx)|$. Without loss of generality, let $\vx\in\Omega_1$, because the case $\vx\in\Omega_2$ can be proved similarly. However, this has already been proved in the proof of Theorem \ref{thm:sep}, where we showed that with probability at least $1-\delta-C_4\exp(-C_5n)$, $f_{\bW(k),\ba}(\vx)\geq 1/4$ for all $\bx\in \Omega_1$.

\subsection{Proof of Theorem \ref{thm:nonsepcali}}
By applying the interpolation inequality, the $L_\infty$ norm of $2\eta - 1 - f_{\bW(k),\ba}$ can be bounded by
\begin{align}\label{eq:inter1}
    \|2\eta - 1-f_{\bW(k),\ba}\|_{\infty} \leq & C_0\|2\eta - 1-f_{\bW(k),\ba}\|_{2}^{\frac{1}{d}}\|2\eta - 1-f_{\bW(k),\ba}\|_{W^{\nu}}^{\frac{d-1}{d}}\nonumber\\
    \leq & C_1\|2\eta - 1-f_{\bW(k),\ba}\|_{2}^{\frac{1}{d}}\|2\eta - 1-f_{\bW(k),\ba}\|_{\mathcal{N}}^{\frac{d-1}{d}}\nonumber\\
    \leq & C_2\|2\eta - 1-f_{\bW(k),\ba}\|_{2}^{\frac{1}{d}}\left(\|2\eta - 1\|_{\mathcal{N}}+\|f_{\bW(k),\ba}\|_{\mathcal{N}}\right)^{\frac{d-1}{d}}\nonumber\\
    \leq & C_3\|2\eta - 1-f_{\bW(k),\ba}\|_{2}^{\frac{1}{d}}\nonumber\\
    \leq & C_4\left(\mathbb{E}_{X\sim P_X}(2\eta(X) - 1-f_{\bW(k),\ba}(X))^2\right)^{\frac{1}{2d}} = O_{\mathbb{P}}(n^{-\frac{1}{4d-2}})
\end{align}
where the second equality is by the equvilance of the Sobolev space $W^\nu$ and the RKHS $\mathcal{N}$; the third inequality is by the triangle inequality; the fourth inequality is by Assumptions \ref{as2} and \ref{as4}; the fifth inequality is because of Assumption \ref{as5}; and the last equality (with big $O$ notation) is because of \Eqref{eq:misratesep1I2}. This finishes the proof.

\subsection{Proof of Lemma \ref{lemma:noisy_tsybakov}}

Let's first consider the simplest $d=1$ case where $\Omega_1=\{\gamma\}$ and $\Omega_2=\{-\gamma\}$. Let $\phi$ denote the standard normal $N(0,1)$ density.
By injecting Gaussian noises $N(0,\upsilon^2)$, the induced conditional probability can be written as 
\[
\tilde{\eta}_\upsilon(x) =\frac{\phi(\frac{x-\gamma}{\upsilon})}{\phi(\frac{x-\gamma}{\upsilon})+\phi(\frac{x+\gamma}{\upsilon})}=\frac{1}{1+\exp(-\frac{2\gamma x}{\upsilon^2})}.
\]
For small enough $1/2>t>0$, direct calculation yields $\{x\in\RR: |2\tilde{\eta}_\upsilon(x) - 1| < t\}=(-x_t,x_t)$ where 
\[
x_t = \frac{\upsilon^2}{2\gamma}\log\rbr{\frac{1+t}{1-t}}\le \frac{2\upsilon^2}{\gamma}t. 
\]
Hence, 
\begin{align*}
    P_{X}(|2\tilde{\eta}_\upsilon(x) - 1| < t) &= P_X(-x_t<x<x_t)\le 2x_t\phi((\gamma+x_t)/\upsilon)\\
    &\le \frac{C\upsilon^2}{\gamma}\exp\left({-\frac{\gamma^2}{2\upsilon^2}}\right) \cdot t.
\end{align*}
In general cases, notice that Tsybakov's noise condition measures the separation between classes. Therefore, the bottleneck for the inequality is where $\Omega_1$ and $\Omega_2$ are the closest, i.e., where margin $2\gamma$ is attained. Let $\bx_+\in \Omega_1$ and $\bx_-\in \Omega_2$ satisfy $\|\bx_+ -\bx_-\|_2=2\gamma$ (which can be attained since $\Omega$ is compact). Consider the delta distribution at $\bx_+$ and $\bx_-$, which is less separated than the original distribution. Then, it reduces to the simplest case.

\subsection{Proof of Theorem \ref{thm:noise}}

A closer look at the proof of Theorem \ref{thm:nonsepthm1} reveals that the convergence rate depends polynomially on the constant in Tsybakov's noise condition. Specifically, it can be checked that $\|\tilde{\eta}_\upsilon\|_{\mathcal{N}}$ converges to infinity and $\mu$ converges to zero polynomially with $\upsilon\to 0$. Under Tsybakov's noise condition, the convergence rate can be obtained via the proof of Theorem \ref{thm:nonsepthm1} as 
\[
L(\hat{f}) - L^* = O_\PP(C^{\frac{1}{\kappa+2}}n^{-\frac{d(\kappa+1)}{(2d-1)(\kappa + 2)}}) = O_\PP\left({\rm poly}\left(\frac{1}{\upsilon}\right)C^{\frac{1}{\kappa+2}}n^{-\frac{d(\kappa+1)}{(2d-1)(\kappa + 2)}}\right) .
\]
In the Gaussian noise injection case, if we choose $\upsilon=\upsilon_n= n^{-1/2}$, applying Lemma \ref{lemma:noisy_tsybakov} yields
\[
L(\hat{f}) - L^* = O_\PP(e^{-n\gamma /6}{\rm poly}\left(n\right))=O_\PP(e^{-n\gamma /7}).
\]

\subsection{Proof of Theorem \ref{prop:labeltr}}
Direct computation implies that
\begin{align*}
    f_i^*(\vx) = \sum_{j=1}^K\eta_j(\vx)v_{ji},
\end{align*}
which implies 
\begin{align}\label{eq:prola1}
    f^*(\vx) = (\vv_1,...,\vv_K)\eta(\vx),
\end{align}
where $v_{ji}$ is the $i$-th element of $\bv_j$.
Let $\bV=(\bv_1,...,\bv_K)$. Multiplying $\bV^\top$ on both sides of \Eqref{eq:prola1} leads to 
\begin{align}\label{eq:prola2}
    \bV^\top f^*(\vx) = & \bV^\top(\vv_1,...,\vv_K)\eta(\vx)= \left(\frac{K}{K-1}\bI-\frac{1}{K-1}\mathbf{1}\mathbf{1}^\top\right)\eta(\vx)\nonumber\\
    = & \frac{K}{K-1}\eta(\vx)-\frac{1}{K-1}\mathbf{1}\mathbf{1}^\top\eta(\vx)\nonumber\\
    = & \frac{K}{K-1}\eta(\vx)-\frac{1}{K-1}\mathbf{1},
\end{align}
where $\mathbf{1}=(1,...,1)^\top$. In \Eqref{eq:prola2}, the second equality is because $\vv_i^\top\vv_j=-1/(K-1)$ if $i\neq j$ and $\vv_i^\top\vv_i=1$; and the last equality is because $\sum_{i=1}^n \eta_i(\vx)=1$. By \Eqref{eq:prola2}, it can be seen that 
\begin{align*}
    \eta_j(\vx) = \frac{(K-1)\vv_j^\top f^*(\vx)+1}{K},
\end{align*}
which finishes the proof.

\section{Proof of Lemmas in the Appendix}
\label{app:prooflemma}
\subsection{Proof of Lemma \ref{lem:condofstein}}

We follow the approach in the proof of Lemma 6.1 and Proposition 6.3 in \cite{steinwart2007fast}. Note that $f_{l, P} = 2p-1$ minimizes $\mathcal{R}_{l, P}$. We first show that for all $f \in \mathcal{F}$ and all $\alpha\geq 0$, 
\begin{align}\label{eq:lem61stein1}
        \mathbb{E}_{X,Y\sim P}(l_1 \circ f - l_1 \circ f_{l,P})^2 
    \leq  C_{\eta,\kappa}(\|f\|_\infty + 1)^{\frac{2\kappa+4\alpha}{\kappa+\alpha}}\|(2\eta-1)^{-1}\|_{q,\infty}^{\frac{\alpha}{\kappa+\alpha}}\mathbb{E}_{X,Y\sim P}(l_1 \circ f - l_1 \circ f_{l,P})^{\frac{\kappa}{\kappa+\alpha}},
\end{align}
where $C_{\eta,\kappa} := \vert|(2\eta - 1)^{-1} \vert|_{\kappa,\infty} + 4$. In particular, one can take $\alpha = 0$ and obtain
\begin{align}\label{eq:lem61stein2}
        \mathbb{E}_{X,Y\sim P}(l_1 \circ f - l_1 \circ f_{l,P})^2 
    \leq  C_{\eta,\kappa}(\|f\|_\infty + 1)^2\mathbb{E}_{X,Y\sim P}(l_1 \circ f - l_1 \circ f_{l,P}).
\end{align}
Clearly, Tsybakov's noise condition implies that $\|(2\eta-1)^{-1}\|_{\kappa,\infty}$ exists. For $\bx \in \Omega$, let $p : = \mathbb{P}(Y=1|\bx)$ and $t : = f(\bx)$. Without loss of generality, let $p>1/2$. Additionally, we denote 
\begin{align}\label{eq:lem1mp1}
v(p, t)= & p\left(l(1, t)-l\left(1, f_{l, P}(\bx)\right)\right)^{2}+(1-p)\left(l(-1, t)-l\left(-1, f_{l, P}(\bx)\right)\right)^{2}, \nonumber\\
m(p, t)=& p\left(l(1, t)-l\left(1, f_{l, P}(\bx)\right)\right)+(1-p)\left(l(-1, t)-l\left(-1, f_{l, P}(\bx)\right)\right). 
\end{align}
Note $f_{l, P} = 2p-1$ implies $l\left(1, f_{l, P}(\bx)\right) = 4(p-1)^2$ and  $l\left(-1, f_{l, P}(\bx)\right) = 4p^2$. Plugging them into \Eqref{eq:lem1mp1}, it can be checked that
\begin{align*}
    m(p, t) =  & 1+t^2 +2(1-2p)t - 4p(1-p) = (1+t-2p)^2,\\
    v(p,t) = & (1+t-2p)^2((t+1)^2 + 12p-4pt-12p^2).
\end{align*}
By taking 
\begin{align}\label{eq:lem1alpha}
    \alpha \ge \frac{\log 4 - \log (12p - 12p^{2} - 4pt +2 - (t - 1)^2)}{\log |2p - 1|},
\end{align}
it can be shown that 
\begin{equation}
\label{eq:ineq1}
    v(p,t)\leq \left( 2t^2 + \frac{4}{|2p-1|^{\alpha}} \right)m(p,t).
\end{equation}
Since 
\begin{align*}
    \frac{\log 4 - \log (12p - 12p^{2} - 4pt +2 - (t - 1)^2)}{\log |2p - 1|}\leq \frac{\log 4 - \log (-\frac{2}{3}t^{2} + 4)}{\log |2p - 1|}\leq 0,
\end{align*}
it suffices to take $\alpha \geq 0$. We further define
\begin{align*}
     g(y, \bx) &:= l(y, f(\bx)) - l(y, f_{l,P}(\bx)),\\
    h_{1}(\bx) &:= \eta(\bx)g(1,\bx) + (1 - \eta(\bx))g(-1,\bx),\\
    h_{2}(\bx) &:= \eta(\bx)g^{2}(1,\bx) + (1 - \eta(\bx))g^{2}(-1,\bx).
\end{align*}

Therefore, \Eqref{eq:ineq1} implies $h_2(\bx)\leq (2\|f\|_\infty^{2} + \frac{4}{|2\eta(\bx)-1|^\alpha})h_1(\bx)$ for all $\bx$ with $\eta(\bx) \neq 1/2$. Hence, we obtain
\begin{align*}
    \mathbb{E}_{X,Y\sim P}g^2 = & \int_{\{\vx||2\eta(\vx)-1|^{-1}< t\}}h_2(\vx)dP_X + \int_{\{\vx|\infty > |2\eta(\vx)-1|^{-1}\geq t\}}h_2(\vx)dP_X\\
    \leq & (2\|f\|_\infty^2 + 4t^\alpha)\int_{\{\vx||2\eta(\vx)-1|^{-1}< t\}}h_1(\vx)dP_X + \int_{\{\vx|\infty > |2\eta(\vx)-1|^{-1}\geq t\}}(\|f\|_\infty + 1)^4dP_X\\
    \leq & 4(\|f\|_\infty^2 + t^\alpha)\mathbb{E}_{X,Y\sim P}g + (\|f\|_\infty + 1)^4\|(2\eta-1)^{-1}\|_{q,\infty}t^{-\kappa}\\
    \leq & 4t^\alpha(\|f\|_\infty + 1)^{2}\mathbb{E}_{X,Y\sim P}g + (\|f\|_\infty + 1)^4\|(2\eta-1)^{-1}\|_{q,\infty}t^{-\kappa}\\
    \leq & 4t^\alpha(\|f\|_\infty + 1)^2\mathbb{E}_{X,Y\sim P}g + (\|f\|_\infty + 1)^4\|(2\eta-1)^{-1}\|_{\kappa,\infty}t^{-\kappa}\\
    \leq & C_{\eta,\kappa}(\|f\|_\infty + 1)^{\frac{2\kappa+4\alpha}{\kappa+\alpha}}\|(2\eta-1)^{-1}\|_{\kappa,\infty}^{\frac{\alpha}{\kappa+\alpha}}\mathbb{E}_{X,Y\sim P}g^{\frac{\kappa}{\kappa+\alpha}},
\end{align*}
where the last equality is implied by taking $t^{\kappa + \alpha} := (\vert|f\vert|_{\infty} + 1)^{2}(\mathbb{E}_{X,Y\sim P} g)^{-1}$. This shows \Eqref{eq:lem61stein1} holds.

Based on \Eqref{eq:lem61stein1}, we can show that Lemma \ref{lem:stein} holds. To see this, let $\hat{C} := (K\gamma + 1)^{(2\kappa+4\alpha)/(\kappa+\alpha)}$ and fix an $f \in \gamma B_{\mathcal{N}}$. The term $\mathbb{E}_{X,Y\sim P}(L_{1} \circ f - L_{1} \circ f_{n})^2$ can be bounded by
\begin{align}\label{eq:lem61stein2}
       &  \mathbb{E}_{X,Y\sim P}(L_{1} \circ f - L_{1} \circ f_{n})^2 \nonumber\\
        \le  &  2\mu^{2}n^{-2} \vert| f \vert|^{4} + 2\mu^{2}n^{-2}\vert| f_{n}\vert|^{4} + 2 \mathbb{E}_{X,Y\sim P}(l_1 \circ f - l_1 \circ f_{n})^{2}\nonumber\\
        \le  & 4 \mathbb{E}_{X,Y\sim P}(l_1 \circ f - l_1 \circ f_{l,P})^{2} + 4 \mathbb{E}_{X,Y\sim P}(l_1 \circ f_{l,P} - l_1 \circ f_{n})^{2} + 2\mu^{2}n^{-2}\vert| f \vert|^{4} + 2\mu^{2}n^{-2} \vert|f_{n}\vert|\nonumber\\
         \le & 8 C_{\eta,\kappa}\hat{C}(\mathbb{E}_{X,Y\sim P}(l_1 \circ f - l_1\circ f_{l,P}) + \mathbb{E}_{X,Y\sim P}(l \circ f_{n} - l \circ f_{l,P}))^{\kappa/(\kappa+\alpha)} + 2\mu^{2}n^{-2}\vert| f\vert|^{4} + 2\mu^{2}\vert|f_{n}\vert|^{4}\nonumber\\
         \le & C\hat{C}(\mathbb{E}_{X,Y\sim P}(l \circ f - l \circ f_{l,P}) + \mathbb{E}_{X,Y\sim P}(l \circ f_{n} - l \circ f_{l,P}) + \mu^{2}n^{-2}\vert| f\vert|^{4} + \mu^{2}n^{-2}\vert|f_{n}\vert|^{4})^{\kappa/(\kappa+\alpha)} \nonumber\\
           \le & C\hat{C}(\mathbb{E}_{X,Y\sim P}(L_{1} \circ f - L_{1} \circ f_{n}) + 2\mathbb{E}_{X,Y\sim P}(l \circ f_{n}  - l \circ f_{l,P})  + 2\mu n^{-1}\vert|f_{n}\vert|^{2})^{\kappa/(\kappa+\alpha)} \nonumber\\
          \le & C\hat{C}(\mathbb{E}_{X,Y\sim P}(L_{1} \circ f - L_{1} \circ f_{n}))^{\kappa/(\kappa+\alpha)} + 2C\hat{C}a^{\kappa/(\kappa+\alpha)}(\mu).
\end{align}
In \Eqref{eq:lem61stein2} the first and second inequalities is because of the Cauchy-Schwarz inequality; the third inequality is because of \Eqref{eq:lem61stein1} and $a^{p} + b^{p} < 2(a + b)^{p}$ for all $a, b \ge 0, 0 < p \le 1$; the fourth inequality follows from $a^{p} + b^{p} < 2(a + b)^{p}$ for all $a, b \ge 0, 0 < p \le 1$; the fifth inequality is because $n^{-1}\mu\vert| f\vert|^{2}\le 1$ and $ n^{-1}\mu\vert|f_n\vert|^{2} \le 1$; the last inequality follows $(a + b)^{p} < a^{p} + b^{p}$ for all $a, b \ge 0, 0 < p \le 1$. This finishes the proof of Lemma \ref{lem:stein}.

\subsection{Proof of Lemma \ref{lem:as3}}
Since there is a positive margin between $\Omega_1$ and $\Omega_2$, we can always find two sets $\tilde{\Omega}_1$ and $\tilde{\Omega}_2$ with infinitely smooth boundaries such that $\Omega_1\subset \tilde{\Omega}_1$, and $\Omega_2\subset \tilde{\Omega}_2$. Then the result follows from the Sobolev extension theorem.

\section{Appendix for Detailed Experiments}
\label{sec:appendix-expriments}

\subsection{Synthetic Data}
\label{app:synthetic}

During the neural network training, we use the popular RMSProp optimizer with the default settings, and select the tuning parameter $\mu$ for SL-ONN + $\ell_2$ and CE-ONN + $\ell_2$ by a validation set. 

\begin{figure}[H]
    \centering
    \includegraphics[width=0.8\textwidth
    ]{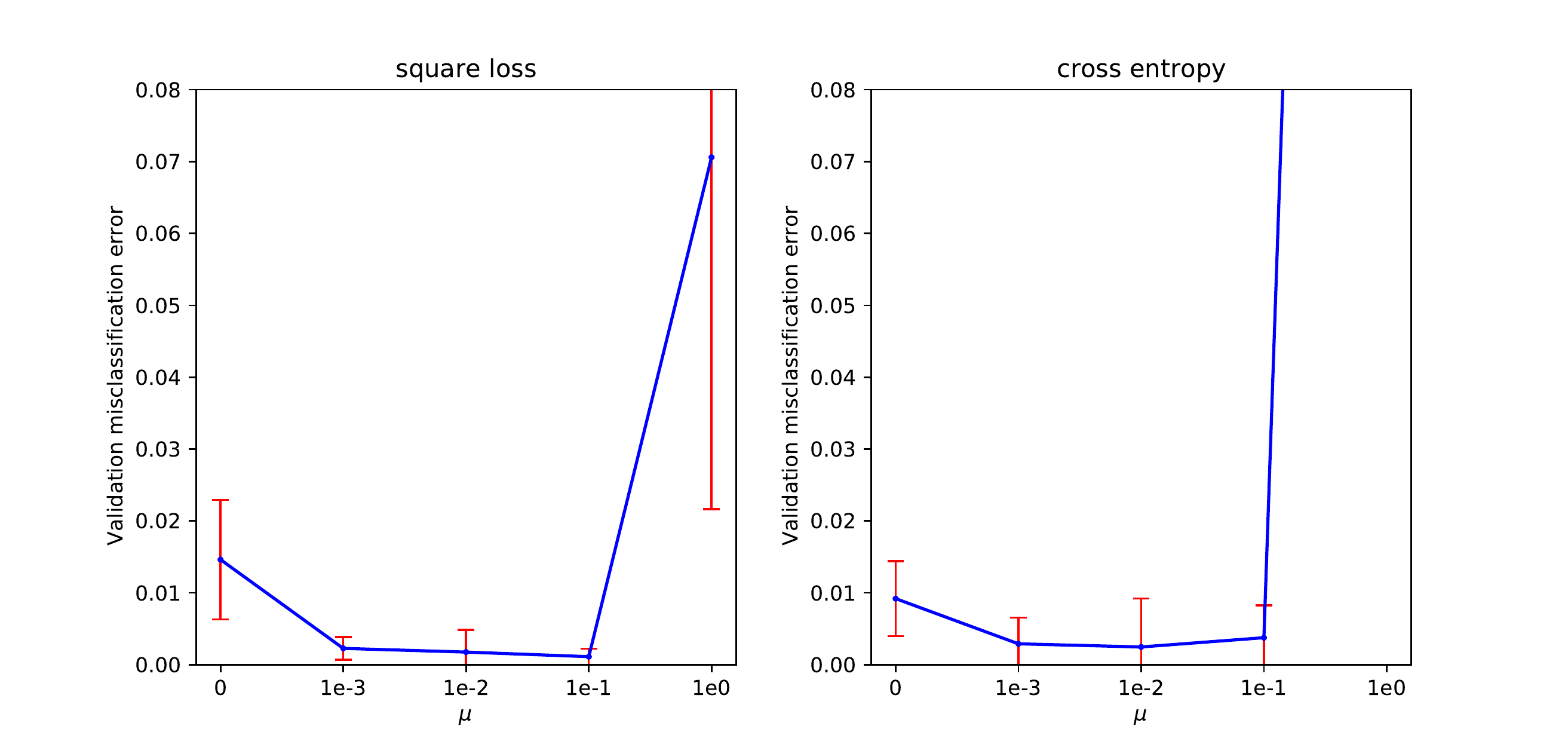}
    \vspace{-3mm}
    \caption{The errorbar plot of validation misclassification rate with respect to different $\mu$ in the separable case.}
    \label{fig:crossval}
    \vspace{-3mm}
\end{figure}

\textbf{Separable case}\quad In the separable case, we consider a two-dimension distribution $P = (\rho \sin \theta + 0.04,  \rho \cos \theta)$ where $\rho = {(\theta/4 \pi)}^{4/5} + \epsilon$ with selected $\theta$ from $(0, 4\pi]$ and $\epsilon \sim \text{unif}([-0.03,0.03])$. We draw 100 positive and 100 negative training samples from $-P$ and $P$, respectively. We select the tuning parameter $\mu$ for SL-ONN + $\ell_2$ and CE-ONN + $\ell_2$ by minimizing the validation misclassification rate, where the candidate set of $\mu$ is $\{0, 0.001, 0.01, 0.1, 1\}$. For each $\mu$, we generate 40 replications to estimate the mean and standard deviation of validation misclassification rate. We observe that SL-ONN + $\ell_2$ and CE-ONN + $\ell_2$ have the least mean and least standard deviation for the validation misclassification rate at $\mu = 0.1$ and $\mu = 0.01$, respectively. The errorbar plot for each $\mu$ is shown in Figure \ref{fig:crossval}. 

\begin{figure}[H]
    \centering
    \subfigure{
      \includegraphics[width=0.32\textwidth]{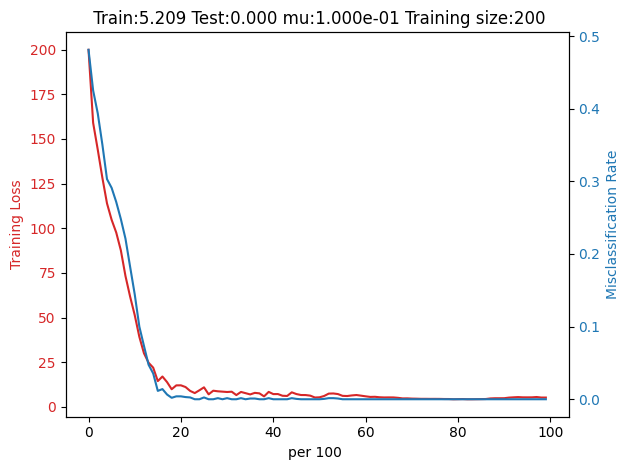}}
    \subfigure{
      \includegraphics[width=0.32\textwidth]{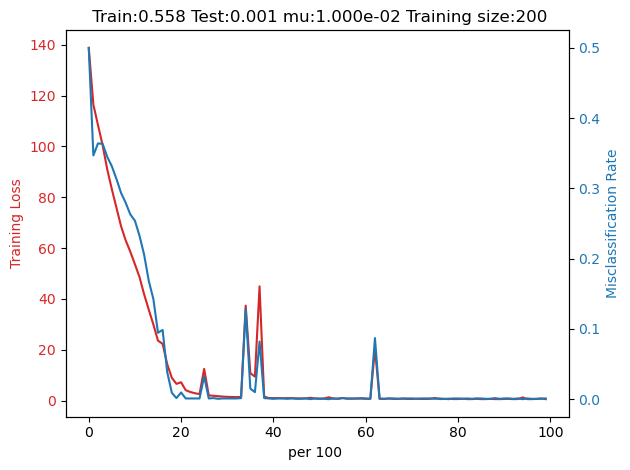}}
    \subfigure{
      \includegraphics[width=0.32\textwidth]{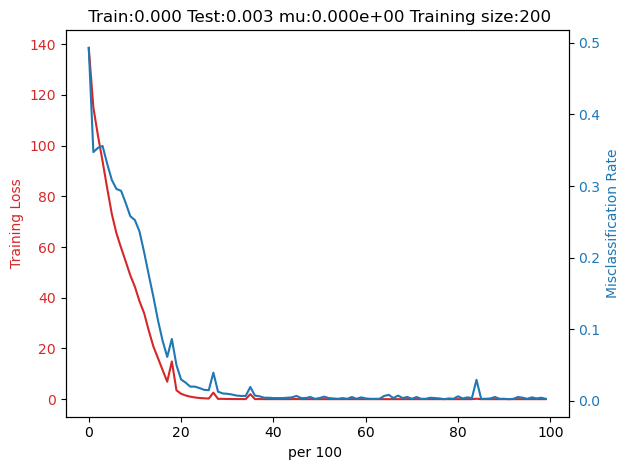}}
    \vspace{-3mm}
    \caption{An instance about the training process of  SL-ONN + $\ell_2$ \figleft , CE-ONN + $\ell_2$ \figcenter~ and CE-ONN \figright. }
    \label{app_fig:sep_training}
    \vspace{-5mm}
\end{figure}

\begin{figure}[htp]
    \centering
    \subfigure{
      \includegraphics[width=1\textwidth]{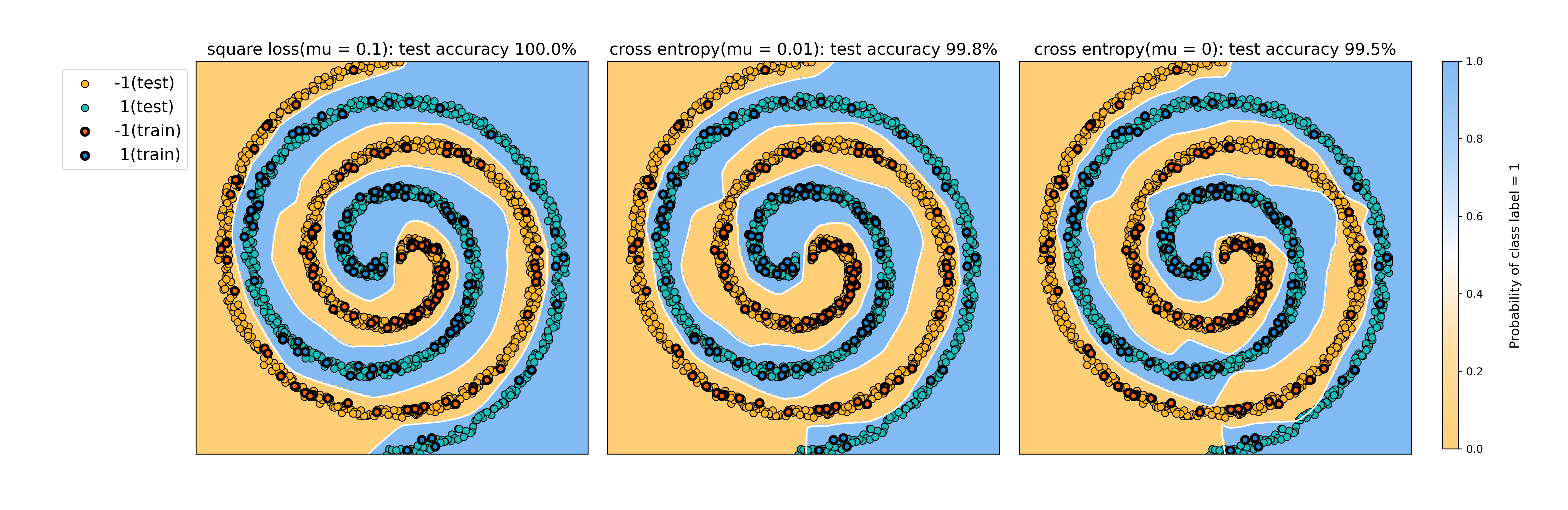}}\vspace{-1cm}
    \subfigure{
      \includegraphics[width=1\textwidth]{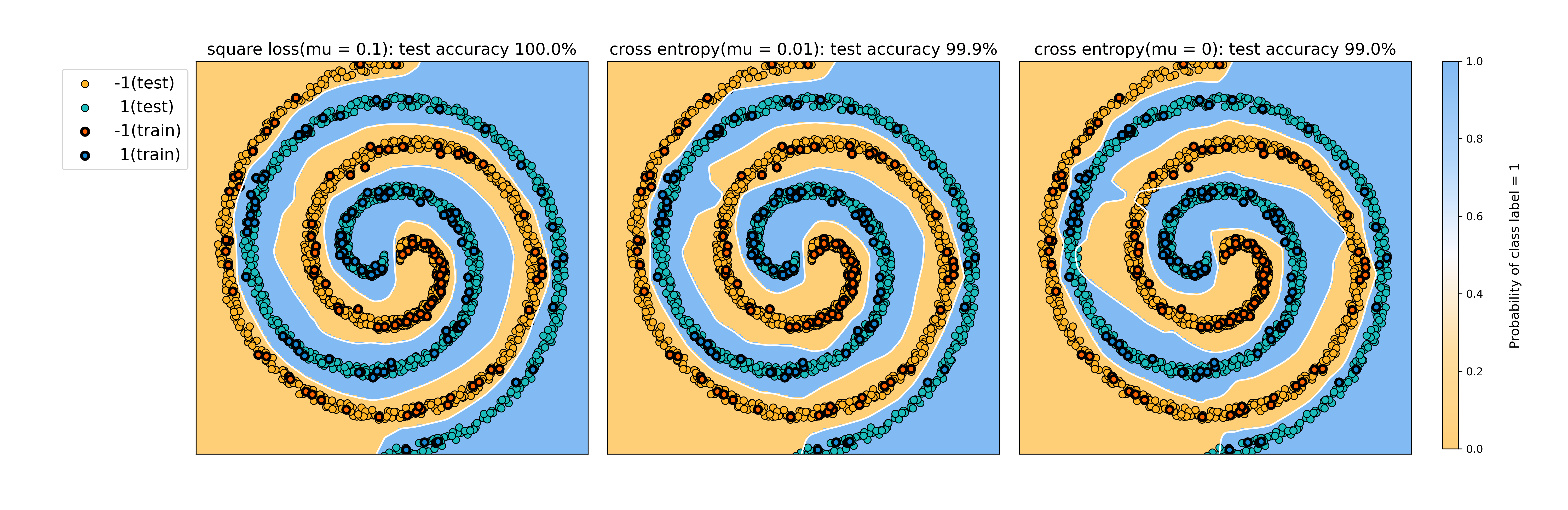}}\vspace{-1cm}
    \subfigure{
      \includegraphics[width=1\textwidth]{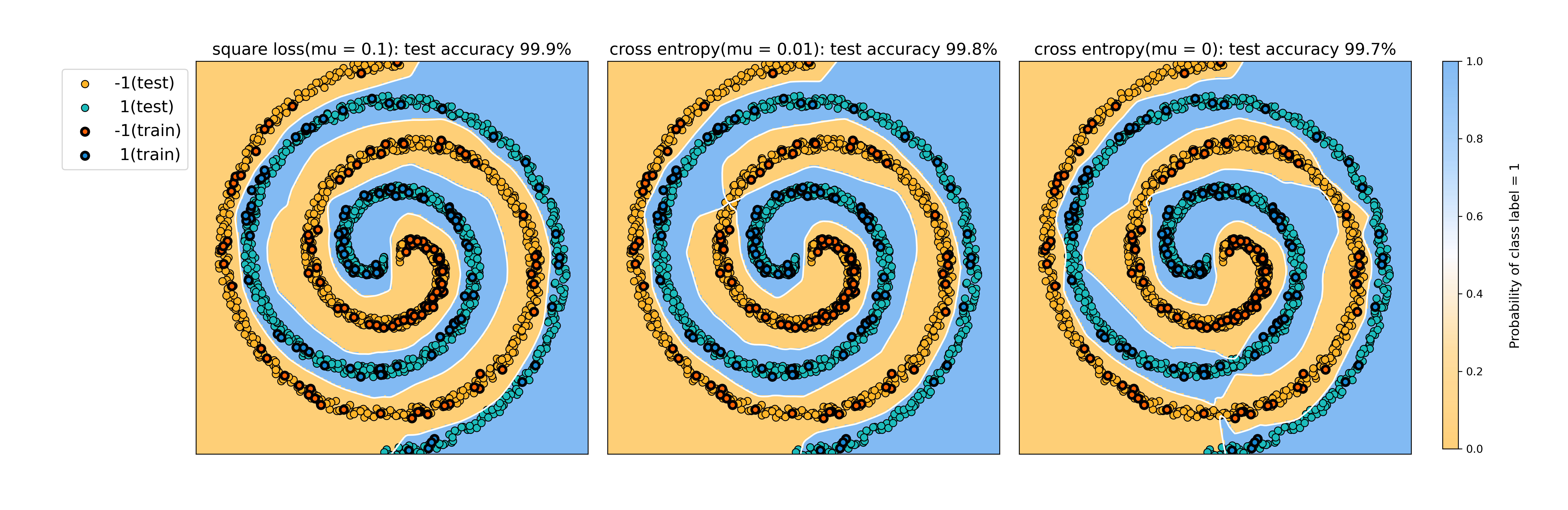}}\vspace{-1cm}
    \subfigure{
      \includegraphics[width=1\textwidth]{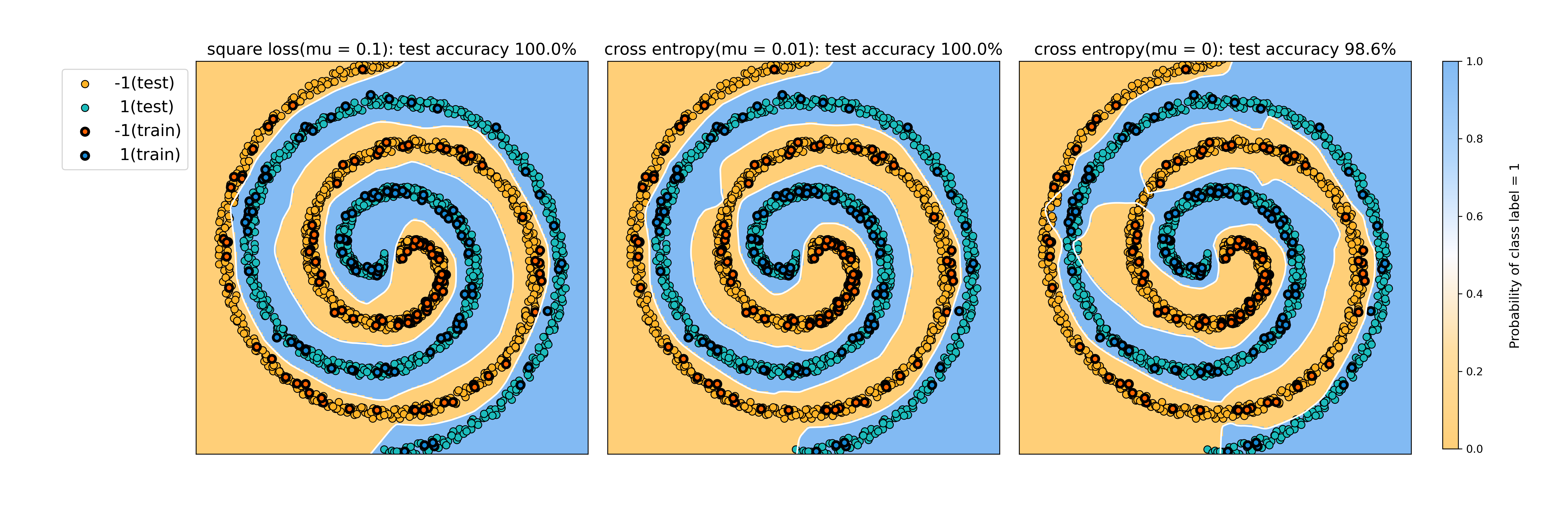}}\vspace{-1cm}
    \subfigure{
      \includegraphics[width=1\textwidth]{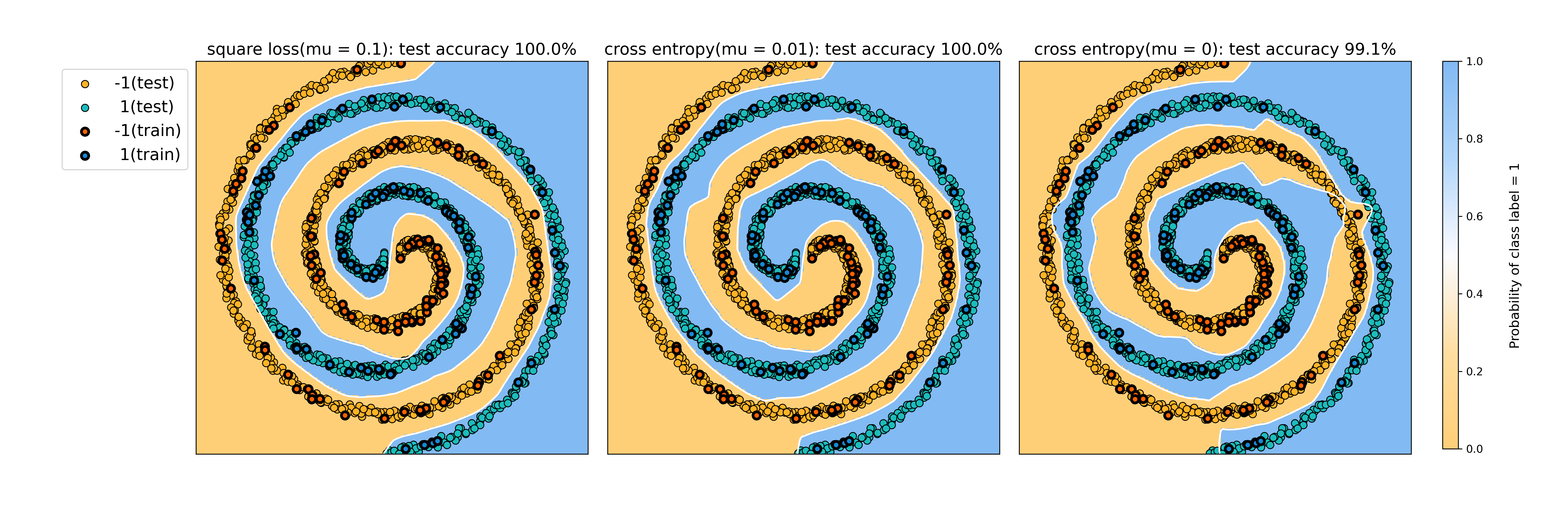}}
    \caption{Five examples of the separable case.}
    \label{app_fig:sep}
\end{figure}

As mentioned in Section \ref{simulation}, we also consider the cross-entropy loss based ONN without $\ell_2$ regularization (CE-ONN). All three models are trained for 10000 iterations and achieve 0\% training misclassification rate. In Figure \ref{app_fig:sep}, we present five more examples about the decision boundary prediction and test accuracy of SL-ONN + $\ell_2$, CE-ONN + $\ell_2$ and CE-ONN. We can find that SL-ONN + $\ell_2$ still beats CE-ONN + $\ell_2$ and CE-ONN in almost all the cases. SL-ONN + $\ell_2$ attains the smallest misclassification rate and depicts the largest margin decision boundary which separates the positive and negative samples best. In addition, we can observe that CE-ONN + $\ell_2$ outperforms CE-ONN in all cases, although the $\ell_2$ regularization term bring some oscillation to the training of CE-ONN + $\ell_2$, as shown in Figure \ref{app_fig:sep_training}. 

\begin{figure}[H]
    \centering
    \includegraphics[width=0.7\textwidth]{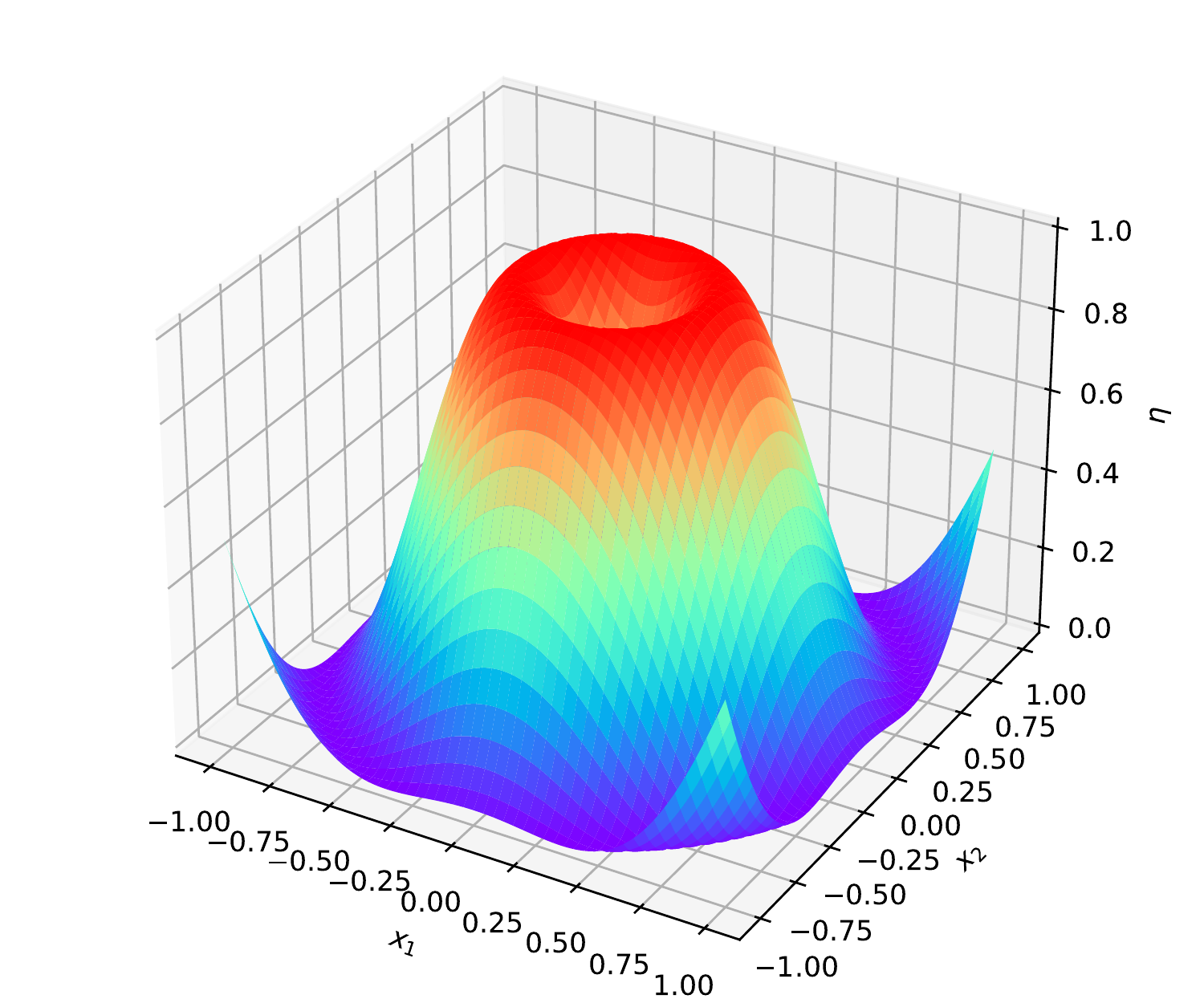}
    \caption{The 3-dimensional plot of $\eta(\bx)$ in the non-separable case.}
    \label{fig:eta}
    \vspace{-5mm}
\end{figure}

\textbf{Non-separable case}\quad
In the non-separable case, the training data points $\vx_{1}, \cdots ,\vx_{n}$ are i.i.d. sampled from unif$([-1, 1]^2)$ and the training labels ${y}_{1},  \cdots ,{y}_{n}$ are sampled according to $Bernoulli(\eta(\vx_i))$, where $\eta(\boldsymbol{x}) = \sin(\sqrt{2}\pi\vert|\boldsymbol{x}\vert|_{2})$, and $n=8000$. The 3-dimensional plot of $\eta(\vx)$ is presented in Figure \ref{fig:eta}. 

\begin{figure}[H]
    \centering
    \includegraphics[width=1\textwidth]{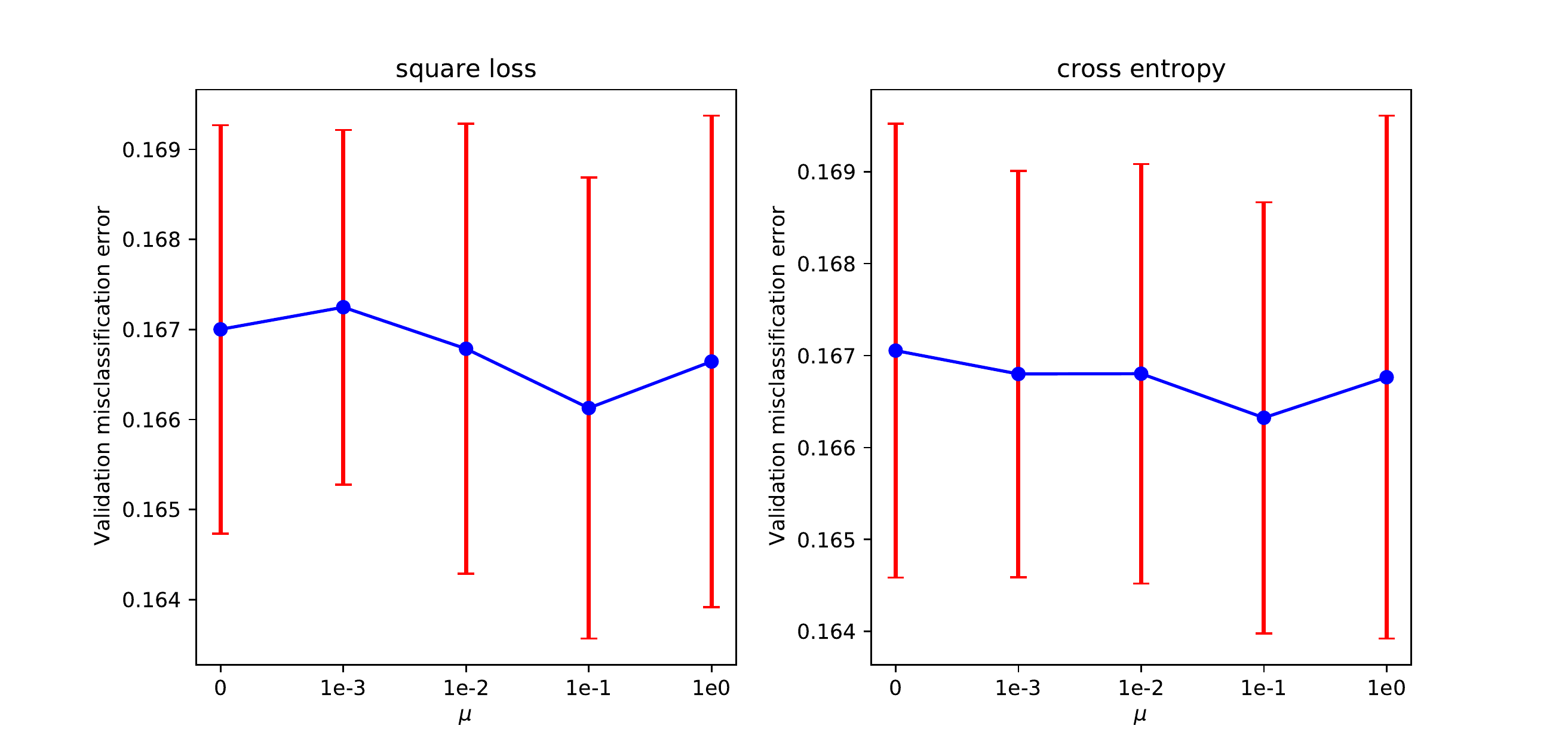}
    \caption{The errorbar plot of validation misclassification rate with respect to different $\mu$ in the non-separable case.}
    \label{fig:crossval_nonsep}
    \vspace{-5mm}
\end{figure}

We select the tuning parameter $\mu$ for SL-ONN + $\ell_2$ and CE-ONN + $\ell_2$ via a validation set, and the candidate set of $\mu$ is $\{0, 0.001, 0.01, 0.1, 1\}$. For each $\mu$, we run 40 replications to estimate the mean and standard deviation of validation misclassification rate. The iteration number of training is 2000. We find SL-ONN + $\ell_2$ and CE-ONN + $\ell_2$ have the smallest mean and standard deviation for the validation misclassification rate at $\mu = 0.1$ and $\mu = 0.1$, respectively. The error bar plot \footnote{In an error bar plot, the center of each plot is the mean, and the upper and lower red dashes denote (mean$+$one standard deviation) and (mean $-$ one standard deviation), respectively.} for $\mu$ equaling to 0, 0.001, 0.01, 0.1 and 1 is shown in Figure \ref{fig:crossval_nonsep}.

\begin{figure}[htbp]
    \centering
    \includegraphics[width=0.36\textwidth]{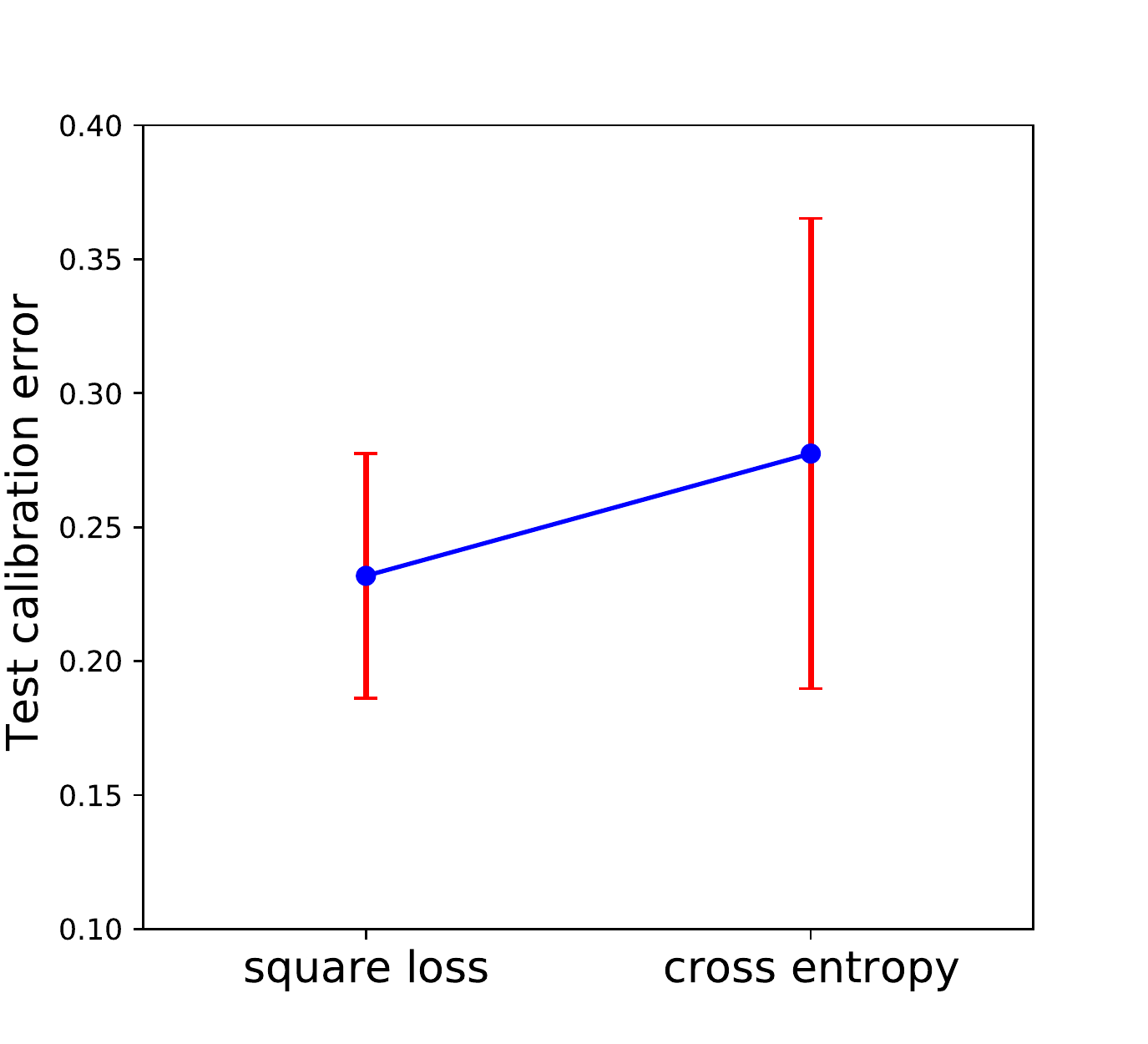}
    \includegraphics[width=0.435\textwidth]{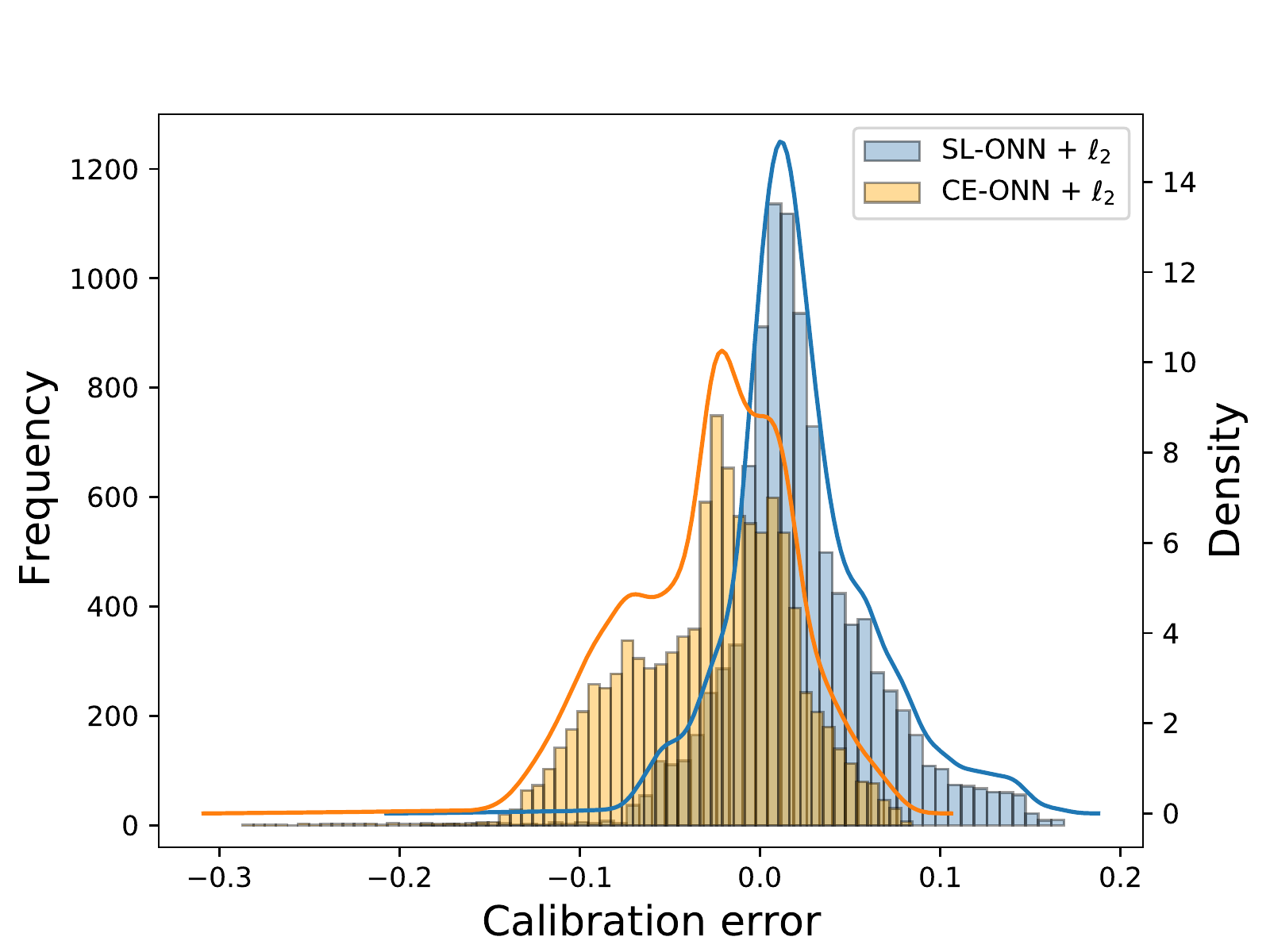}
    \vspace{-1mm}
    \caption{(Left) The error bar plot of test calibration errors for the 40 replicated runs. (Right) The histogram and kernel density estimation of test calibration errors in one instance from 40 replications. In this instance, $\vert|(\hat{f}_{l2} + 1)/2 - \eta\vert|_{L_\infty}$ = 0.188,  $\vert|(\hat{f}_{ce} + 1)/2 - \eta\vert|_{L_\infty}$ = 0.287, and the test misclassification rate for $\hat{f}_{l2}$ and $\hat{f}_{ce}$ are 0.167 and 0.164, respectively. 
    \label{fig:non_sep}}
    \vspace{-5mm}
\end{figure}

The calibration error results are presented in Figure \ref{fig:non_sep}. The error bar plot of the test calibration error shows that $\hat{f}_{l2}$ has the smaller mean and standard deviation than $\hat{f}_{ce}$.

\subsection{Real Data}
\label{app:realdata}
\paragraph{Data and network architecture} 
We use the popular CIFAR-10 and CIFAR-100 datasets, with training and testing split of 50000 and 10000. 
The data loader is from \verb|torch.utils.data|.
As typically employed in practice, our training includes data augmentations, a composition of random crop and horizontal flip. 
We trained two types of neural networks, ResNet \citep{he2016deep} and Wide ResNet \citep{zagoruyko2016wide}. To be more specific, we used the default ResNet-18 and ResNet-50 for CIFAR-10 and CIFAR-100 respectively, and the default WRN-16-10 for both CIFAR-10 and CIFAR-100. All experiments are run in PyTorch version 1.9.0 and cuda 10.2.

\paragraph{Training details} 
The training algorithm is the default SGD with momentum ($0.9$) and weight decay ($0.0005$). 
The learning rate scheduler is the \verb|StepLR()| from \verb|torch.optim.lr_scheduler| with step size 50. 
In our experiment, the only parameters that we tuned are the learning rate (lr) and batch size (bs), with only two options, (lr=0.01, bs=32) and (lr=0.1, bs=128). 
We find that (lr=0.01, bs=32) performs better for most cases except for square loss trained WRN-16-10 on CIFAR-100, where the average accuracy for (lr=0.01, bs=32) is 77.96\%, around 1.5\% less than that for (lr=0.1, bs=128). Meanwhile, for cross-entropy trained WRN-16-10, (lr=0.1, bs=128) yields an average accuracy of 76.83\%, around 1\% less than that for (lr=0.01, bs=32). 
The two training settings perform quite comparable for WRN-16-10 on CIFAR-10. For consistency, we stick with (lr=0.01, bs=32) in this case.

\paragraph{Adversarial robustness}
For square loss, training deep classifiers is the same as regression. 
When attacking classifiers trained with square loss, the default way of constructing adversarial examples doesn't work well. To be more specific, for a correctly classified training image $(\bx, y)$, the adversarial examples are typically generated by 
\[\max_{\|\bm{\delta}\|_\infty = \alpha} L(f(\bx+\bm{\delta}), y).
\] 
Such an attacking scheme works fine for cross-entropy, where 
\[
L(f(\bx),y) = -\log(\mathrm{softmax}(f(\bx))) = -\log\rbr{\frac{\exp(f_y(\bx))}{\sum_{k\ne y}\exp(f_{k}(\bx))}},
\]
but is problematic for regression losses such as square loss. 
The fundamental reason lies in Proposition \ref{prop:labeltr} and its proof. 
Recall that the conditional probability for square loss consists of projections of the classifier outputs to all the simplex vertices, some of which are sure to be non-positive. The sum of the class probabilities from  \Eqref{eqn:calibration} is always 1 but unlike that from softmax function, the summand can be negative. 
By maximizing the square loss, the resulting ``adversarial" image can stay the same class but more confidently. 
To illustrate, if $f(\bx)=\bv_y$, the predicted confidence for label $y$ will be 100\% and 0 for other classes. The ``adversarial" image may be such that $f(\bx+\delta) = 2 \bv_y$, where the predicted label remains unchanged but with an updated confidence of $2-1/K$ for label $y$ and $(1/K-1)/(K-1)<0$ for all other classes. This is obviously not a successful attack. 

To this end, we devise a special attacking scheme for classifier trained with square loss and simplex coding. The key idea is to choose attack directions tangent to the sphere inscribed by the simplex. Instead of $$L(f(\bx),y) = \|f(\bx)-\bv_y\|_2^2,$$ we choose $$L(f(\bx),y)= \theta(f(\bx), \bv_y),$$ where $\theta(\bv_1, \bv_2)$ denotes the cosine similarity between $\bv_1$ and $\bv_2$. We refer to this attack as angle attack.

Empirically, we found our angle attack to significantly outperform the naive attack by maximizing the square loss. 
For square loss, let the predicted probabilities from \Eqref{eqn:calibration} be $\hat{\bp}$.
Similar to cross entropy, we have also tried two cases of $L(f(\bx),y)$, which corresponds to 
$$L_1(f(\bx),y) = -\log(\mathrm{softmax}(\hat{p}_y(\bx))) \quad \mbox{and} \quad L_2(f(\bx),y) = -\log(\hat{p}_y(\bx)).$$
Interestingly for PGD-100, $L_1$ performs the best, beating angle attack for the majority cases, except for attacking WRN-16-10 on CIFAR-100 with strength 2/255. 
The reported adversarial accuracy for square loss trained classifiers in Table \ref{table1} is by $L_1(f(\bx),y)=-\log(\mathrm{softmax}(\hat{p}_y(\bx)))$. 

The PGD attack results may be further improved for square loss. 
Nonetheless, the AutoAttack still provides convincing results, as it includes both white-box and black-box attacks. We used the standard version which includes 4 types of attacks, APGD-CE, APGD-DLR, FAB and Square Attack as in \cite{croce2020reliable}.

\paragraph{Robustness to Gaussian Noise}
To make the robustness evaluation more comprehensive, beyond the adversarial robustness, we also investigate the classifier's robustness to Gaussian noise injections. With the image pixels' value normalized to 0 and 1, we consider injecting Gaussian noises to test images and report the test accuracy. The noise standard deviation ranges from 0.1 to 0.4. The test accuracy results for both CIFAR-10 and CIFAR-100 are listed in Table \ref{table:gaussian}.

\input{tables/table_add2}

\paragraph{Simplex coding vs. one-hot coding}
The one-hot coding is the usual choice for applying square loss to classification. 
However, it is empirically observed to struggle when the number of classes are large. For a single training data point $\bx$ and label $k$, \citet{hui2020evaluation} proposed to modify the training objective from the typical $(f_k(\vx)-1)^2 + \sum_{i\ne k} f_i(\vx)^2$ to $J\cdot (f_k(\vx)-M)^2 + \sum_{i\ne k} f_i(\vx)^2$, where $J, M$ are hyperparameters to make $f_k$ more prominent in the loss. 
Similar modification is also proposed in \cite{demirkaya2020exploring}. 
The scaling trick involves two hyperparameters, which can be hard to tune. 
We evaluate the two coding schemes in our experiment setting and the results are summarized in Table \ref{table2}.
The test accuracy for scaled one-hot coding scheme performs comparably for ResNet-18 on CIFAR-10 and ResNet-50 on CIFAR-100. For WRN-16-10, the simplex coding performs better. 

\input{tables/table2}

\end{document}

%% file: math_commands.tex
\usepackage{amsmath,amsfonts,bm}

\newcommand{\figleft}{{\em (Left)}}
\newcommand{\figcenter}{{\em (Center)}}
\newcommand{\figright}{{\em (Right)}}

\def\eqref#1{equation~\ref{#1}}
\def\Eqref#1{Equation~\ref{#1}}

\def\1{\bm{1}}

\def\vv{{\bm{v}}}

\def\vx{{\bm{x}}}

\DeclareMathAlphabet{\mathsfit}{\encodingdefault}{\sfdefault}{m}{sl}
\SetMathAlphabet{\mathsfit}{bold}{\encodingdefault}{\sfdefault}{bx}{n}



%% file: tables/table1.tex
\begin{table}[!ht]
\caption{Test accuracy on CIFAR datasets. Average accuracy larger than 0 but less than 0.1 is denoted as 0$^*$ without standard deviation.}
\vspace*{0.03in}
\label{table1}
\setlength{\tabcolsep}{6pt}
\centering
\Large
\begin{spacing}{1.3}
\resizebox{\linewidth}{!}{
\begin{tabular}{|c|c|c|c|c|c|c|c|c|c|}
\hline
{\multirow{2}{*}{Dataset}} &
{\multirow{2}{*}{Network}} &
{\multirow{2}{*}{Loss}} &
{\multirow{2}{*}{Clean acc \%}} &\multicolumn{3}{c|}{\emph{PGD-100 ($l_\infty$-strength)}}&\multicolumn{3}{c|}{\emph{AutoAttack ($l_\infty$-strength)}} \\\cline{5-10}
 & &  & & $  {2}/{255}$ & $  4/255$ & $  8/255$ & $  2/255$ & $  4/255$ & $  8/255$\\
\specialrule{0em}{0pt}{0pt}
\hline
{\multirow{4}{*}{CIFAR-10}} & \multirow{2}{*}{ResNet-18} & CE & \bf{95.15 (0.11)}  & ~~8.81 (1.61) &~~0.65 (0.24) & 0& 2.74 (0.09)& $0$ &0 \\ \cline{3-10}
& & SL & 95.04 (0.07)& \bf{30.53 (0.92)} & ~~\bf{6.64 (0.67)} & \bf{0.86 (0.24)} & \bf{4.10 (0.50)} & \bf{~~0$^*$} & 0\\ \cline{2-10}
&  \multirow{2}{*}{WRN-16-10} & CE & 93.94 (0.16) & ~~1.04 (0.10)& 0& 0& 0.33 (0.06)& 0& 0 \\ \cline{3-10}
& & SL & \bf{95.02 (0.11)} & \bf{37.47 (0.61)} & \bf{23.16 (1.28)} & \bf{7.88 (0.72)} & \bf{5.37 (0.50)} & ~~\bf{0$^*$} & 0\\ \hline
{\multirow{4}{*}{CIFAR-100}} & \multirow{2}{*}{ResNet-50} & CE & \bf{79.82 (0.14)} & ~~2.31 (0.07) & ~~0$^*$ & 0& 0.99 (0.10) & ~~0$^*$ & 0\\ \cline{3-10}
& & SL & 78.91 (0.14) & \bf{13.76 (1.30)} & ~~\bf{4.63 (1.20)} & \bf{1.21 (0.80)} & \bf{3.67 (0.60)} & \bf{0.16 (0.05)} & 0 \\ \cline{2-10}
&  \multirow{2}{*}{WRN-16-10} & CE & 77.89 (0.21) & ~~0.83 (0.07) & ~~0$^*$ &0 & 0.42 (0.07) & 0&0 \\ \cline{3-10}
& & SL & \bf{79.65 (0.15)} & ~~\bf{6.48 (0.40)} & ~~\bf{0.42 (0.04)} & ~~\bf{0$^*$} & \bf{2.73 (0.20)} & ~~\bf{0$^*$} &0 \\\hline
\end{tabular}}
\end{spacing}

\vspace{-7mm}
\end{table}

%% file: tables/table_add1.tex
\begin{table}[H]
\caption{Performance on CIFAR-10 dataset for ResNet-18 under standard PGD adversarial training. }
\vspace*{-0.05in}
\label{table_adv}
\setlength{\tabcolsep}{6pt}
\begin{center}\scalebox{0.8}{
\resizebox{\linewidth}{!}{
\begin{tabular}{|c|c|c|c|c|c|}
\hline
CIFAR10 & Loss & Acc (\%) &  PGD steps & Strength($l_{\infty}$) & Autoattack \\ \hline
\multirow{2}{*}{ResNet-18} & \multirow{2}{*}{CE} & 86.87  &  3 & 8/255 & 37.08\\ \cline{3-3} \cline{4-6}
& & 84.50 &  7 & 8/255 & 41.88\\ \hline
\multirow{2}{*}{ResNet-18} & \multirow{2}{*}{SL} & \bf{87.31}  &  3 & 8/255 & \bf{40.46}\\ \cline{3-3} \cline{4-6}
& & \bf{84.52} &  7 & 8/255 & \bf{44.76}\\ \hline
\end{tabular}}}
\end{center}
\vspace{-7mm}
\end{table}

%% file: tables/table_add2.tex
\begin{table}[ht]
\caption{Black-box Gaussian noise robustness results. The reported accuracy is the average of 5 replications. }
\vspace*{-0.05in}
\label{table:gaussian}
\setlength{\tabcolsep}{6pt}
\centering
\begin{spacing}{1.2}
\scalebox{0.75}{
\resizebox{\linewidth}{!}{
\begin{tabular}{|c|c|c|c|c|c|c|c|}
\hline
{\multirow{2}{*}{Dataset}} &
{\multirow{2}{*}{Network}} &
{\multirow{2}{*}{Loss}} &
\multicolumn{5}{c|}{\emph{Gaussian noise standard deviation}}\\\cline{4-8}
 & &  &  0.00 & 0.10 & 0.20 & 0.30 & 0.40 \\ \hline
{\multirow{4}{*}{CIFAR-10}} & \multirow{2}{*}{ResNet-18} & SL & 95.04  & \bf{90.07} & \bf{70.16}  & \bf{42.13} & \bf{25.38}\\ \cline{3-8}
& & CE & \bf{95.15} & \bf{90.03} & 69.71 & 41.08 & 24.66 \\ \cline{2-8}
& \multirow{2}{*}{WRN-16-10} & SL &  \bf{95.02} & \bf{88.49} & \bf{60.91} & \bf{35.78} & \bf{24.04} \\ \cline{3-8}
& & CE & 93.94 & 84.78 & 56.63 & 33.70 & 22.41 \\ \hline
{\multirow{4}{*}{CIFAR-100}} & \multirow{2}{*}{ResNet-50} & SL & \bf{78.91} & \bf{63.06} &  \bf{36.64} & \bf{17.78} & ~~\bf{9.47}\\ \cline{3-8}
& & CE & 79.82 & 62.72 & 34.42 & 16.69 & ~~9.11 \\ \cline{2-8}
&  \multirow{2}{*}{WRN-16-10} & SL & \bf{79.65} & \bf{62.01} & \bf{30.69} & \bf{15.11} & ~~\bf{8.88} \\ \cline{3-8}
& & CE & 77.89 & 60.14 & 26.47 & 10.26 & ~~5.57\\\hline
\end{tabular}}
 }
\end{spacing}
\vspace{-3mm}
\end{table}

%% file: tables/table2.tex
\begin{table}[t]
\caption{Test accuracy for square loss with one-hot coding (scaled) (OC) vs. simplex coding (SC).
Accuracy with an asteroid sign ($^*$) denotes cases where the training accuracy doesn't overfit after 200 training epochs.}
\vspace*{-0.05in}
\label{table2}
\setlength{\tabcolsep}{6pt}
\begin{center}
\begin{spacing}{1.2}
\scalebox{1}{
\resizebox{\linewidth}{!}{
\begin{tabular}{|c|c|c|c|c|c|}
\hline
Dataset & Network & One-hot scaling & SGD parameters & OC clean acc(\%) & SC clean acc(\%)\\ \hline
{\multirow{4}{*}{CIFAR-10}} & \multirow{2}{*}{ResNet-18} & \multirow{2}{*}{k=1, M=1} & lr=0.01, bs=32  & 94.95 & 95.04\\ \cline{4-6}
& & & lr=0.1, bs=128 & 10$^{*}~~~$ & 10$^{*}~~~$\\ \cline{2-6}
& \multirow{2}{*}{WRN-16-10} & \multirow{2}{*}{k=1,M =1} & lr=0.01, bs=32  & ~~89.75$^{*}$ & 95.02\\ \cline{4-6}
& & & lr=0.1, bs=128 & ~~88.43$^*$ & 95.03 \\ \hline
{\multirow{4}{*}{CIFAR-100}} & \multirow{2}{*}{ResNet-50} & \multirow{2}{*}{k=5, M=15} & lr=0.01, bs=32  & 79.06 & 78.91\\ \cline{4-6}
& & & lr=0.1, bs=128 & $1^{*}~$ & 1$^*~$\\ \cline{2-6}
& \multirow{2}{*}{WRN-16-10} & \multirow{2}{*}{k=5, M=15} & lr=0.01, bs=32  & 78.42 & 78.06\\ \cline{4-6}
& & & lr=0.1, bs=128 & 78.39 & 79.65\\ \hline
\end{tabular}}}
\end{spacing}
\end{center}
\vspace{-8mm}
\end{table}